%% file: PaperForArxiv.tex
\documentclass[10pt,twocolumn,letterpaper]{article}

\usepackage[pagenumbers]{cvpr} 

\usepackage[table]{xcolor}

\usepackage{subcaption}
\usepackage[font=small,labelfont=bf]{caption}

\usepackage{graphicx}
\usepackage{amsmath}
\usepackage{amssymb}
\usepackage{booktabs}
\usepackage{arydshln}
\usepackage[accsupp]{axessibility} 

\usepackage[export]{adjustbox}
\usepackage[toc,page]{appendix}
\usepackage{indentfirst}

\definecolor{DarkGreen}{rgb}{0.0, 0.6, 0.0}

\definecolor{DarkBlue}{rgb}{0.0, 0.0, 0.8}

\definecolor{DarkOrange}{rgb}{0.9, 0.5, 0.0}

\definecolor{DarkRed}{rgb}{0.9, 0.2, 0.2}

\definecolor{DarkViolet}{rgb}{0.5, 0.2, 0.5}

\newcommand{\ours}{\mbox{FSNet}\xspace}
\newcommand{\maa}{MAA@10\textdegree\xspace}
\newcommand{\mediandeg}{Median (\textdegree)\xspace}
\newcommand{\fmat}{\mbox{fundamental matrix}\xspace}
\newcommand{\fmats}{\mbox{fundamental matrices}\xspace}

\usepackage[pagebackref,breaklinks,colorlinks]{hyperref}

\usepackage[capitalize]{cleveref}
\crefname{section}{Sec.}{Secs.}
\Crefname{section}{Section}{Sections}
\Crefname{table}{Table}{Tables}
\crefname{table}{Tab.}{Tabs.}

\def\cvprPaperID{4258} 
\def\confName{CVPR}
\def\confYear{2023}

\begin{document}

\title{Two-view Geometry Scoring Without Correspondences}

\author{
Axel Barroso-Laguna$^{1}$\hspace{30pt}
Eric Brachmann$^{1}$\hspace{30pt}
Victor Adrian Prisacariu$^{1, 2}$\hspace{30pt}\\
Gabriel Brostow$^{1, 3}$\hspace{30pt}
Daniyar Turmukhambetov$^{1}$\\
{\normalsize $^1$Niantic \hspace{30pt} $^2$University of Oxford \hspace{30pt} $^3$University College London}\\
{\normalsize \href{https:///www.github.com/nianticlabs/scoring-without-correspondences}{www.github.com/nianticlabs/scoring-without-correspondences}}
}

\maketitle

\begin{abstract}
Camera pose estimation for two-view geometry traditionally relies on RANSAC. Normally, a multitude of image correspondences leads to a pool of proposed hypotheses, which are then scored to find a winning model. The inlier count is generally regarded as a reliable indicator of ``consensus''. We examine this scoring heuristic, and find that it favors disappointing models under certain circumstances.

As a remedy, we propose the Fundamental Scoring Network (\ours), which infers a score for a pair of overlapping images and any proposed fundamental matrix. It does not rely on sparse correspondences, but rather embodies a two-view geometry model through an epipolar attention mechanism that predicts the pose error of the two images. \ours can be incorporated into traditional RANSAC loops. We evaluate \ours on fundamental and essential matrix estimation on indoor and outdoor datasets, and establish that \ours can successfully identify good poses for pairs of images with 
few or unreliable correspondences. 
Besides, we show that naively combining \ours with MAGSAC++ scoring approach achieves state of the art results.
\end{abstract}

%
%

\section{Introduction}\label{sec:intro}

How to determine the relative camera pose between two images is one of the cornerstone challenges in computer vision. Accurate camera poses underpin numerous pipelines such as Structure-from-Motion, odometry, SLAM, and visual relocalization, among others~\cite{torr1993outlier, matas2004robust, sattler2016efficient, schonberger2016structure, ghosh2016survey}.
Much of the time, an accurate fundamental matrix can be estimated by existing means, but the failures are prevalent enough to hurt real-world tasks, and are hard to anticipate~\cite{fan2022instability}. Where are the mistakes coming from?

Traditional approaches first detect then describe a set of interest points in each image, and establish correspondences between the two sets while  possibly filtering them, \eg, checking for mutual nearest neighbors or applying Lowe's ratio test~\cite{lowe1999object}. Then, random subsets of correspondences are sampled and a 5-point or 7-point algorithm is used to estimate many essential or fundamental matrix hypotheses, respectively, (\ie, two-view geometry models). 
A RANSAC~\cite{fischler1981random} loop iterates over the generated hypotheses, and ranks them. Conventionally, the ranking is  scored by counting inliers, \ie the number of correspondences within a threshold of that two-view geometry hypothesis. 
Finally, the top-ranked hypothesis is further refined by using all inlier correspondences.


\input{figures/teaser_fig}
As the research in robust model estimation advances~\cite{torr2000mlesac,chum2005matching,barath2020magsac++, raguram2012usac,cavalli2022nefsac,barath2022learning}, the different stages of the pipeline are being revisited, \eg, local feature detection and description is learned with neural networks, outlier correspondences are filtered with learned models, hypotheses are sampled more efficiently, or the inlier threshold is optimized.
Although the latest matching pipelines produce very accurate and robust correspondences \cite{detone2018superpoint, sarlin2020superglue, sun2021loftr}, correspondence-based scoring methods are still sensitive to the ratio of inliers, number of correspondences, or the accuracy of the keypoints \cite{fan2022instability, brachmann2019neural, barath2020magsac++, barath2022learning}. 
Incorrect two-view geometry estimation can lead to invalid merges in 3D reconstruction models \cite{schonberger2016structure}, bad localization services \cite{arnold2022map}, or more expensive steps when finding outliers in pose graphs \cite{chen2021learning}. 

A second family of approaches emerged in recent years, where a neural network is trained to directly regress two-view geometry from the input images~\cite{ranftl2018deep, poursaeed2018deep, zhou2020learn, arnold2022map}. Thus, such approaches replace all the components of the RANSAC pipeline. This can be a viable approach when two views are extremely difficult, even when they do not overlap \cite{cai2021extreme}. However, challenging scenarios, \eg, wide-baseline, or large illumination changes, can lead to incorrect predictions~\cite{jin2021image}.
Typically, poses directly regressed this way have fewer catastrophic relative pose predictions, but they have difficulty in estimating precise geometry~\cite{arnold2022map}. On the other hand, correspondence-based hypotheses can be very precise, if estimated correspondences are of sufficiently high quality. Our approach uses correspondences to generate model hypotheses, but does not use correspondences to score them during the RANSAC loop.

We propose a fundamental matrix scoring network that leverages epipolar geometry to compare features of the images in a dense manner. We refer to our method as the Fundamental Scoring Network, or \ours for short. Inspired by the success of Vision Transformers~\cite{vaswani2017attention}, and detector-free matchers \cite{sun2021loftr, jiang2021cotr}, we define an architecture that incorporates the epipolar geometry into an attention layer, 
and hence the quality of the \fmat hypothesis conditions the coherence of the computed features. Figure~\ref{fig:teaser} shows an example where correspondences are highly populated by outliers. However, there are still enough inliers to generate a good \fmat, and \ours was able to select it from the hypothesis pool. 


Our contributions are 1) an analysis of the causes of scoring failures, as well as more insights into the traditional RANSAC approach of relative pose estimation; 2) \ours, a network that predicts angular translation and rotation errors for a given image pair and a fundamental matrix hypothesis; 3) an image order-invariant design of \ours that outputs the same values for (Image A, Image B, $F$) and (Image B, Image A, $F^T$) inputs; 4) a solution that can be combined with state-of-the-art methods to cope with current failure cases. 


%
%

\section{Related Work}
\label{sec:related_work}

\noindent
\textbf{Establishing correspondences.}
Previous correspondence estimation built around SIFT~\cite{lowe1999object} has largely been superseded by learned methods. 
Keypoint detectors \cite{verdie2015tilde, barroso2019key}, patch-based descriptors \cite{tian2019sosnet, tian2020hynet}, joint detector-descriptors \cite{detone2018superpoint, revaud2019r2d2, dusmanu2019d2, barroso2020hdd}, or shape estimators \cite{yi2016learning, mishkin2018repeatability, barroso2022scalenet} are some of the steps that have benefited from data-driven techniques. To find correspondences, the mutual nearest neighbors and Lowe's ratio test~\cite{lowe1999object} approach has also been revisited, and learned matchers have pushed forward the matching capability of feature extractors \cite{sarlin2020superglue}. Complementary to matchers, additional filtering methods learn to detect and reject outlier correspondences \cite{yi2018learning, zhao2019nm, zhang2019learning, sun2020acne, cavalli2020adalam}.
Once correspondences are established, the RANSAC loop finds the best hypothesis among the pool. Multiple works aim at sampling an all-inlier correspondence minimal-set sooner than RANSAC \cite{torr2000mlesac, chum2005matching, torr2002napsac, brachmann2019neural, tong2021deep}. Combining that with early termination techniques and detection of degenerate configurations \cite{chum2005two, ivashechkin2021vsac, cavalli2022nefsac} can significantly improve the results and run-time.
More recently, alternative methods have been proposed to improve upon the classical detect-then-describe approach, \eg, detector-free matchers \cite{sun2021loftr, jiang2021cotr, truong2021learning}, or direct relative-pose regressor networks \cite{zhou2020learn,arnold2022map,ranftl2018deep, Rockwell2022, cai2021extreme}.

\noindent
\textbf{Model quality.} Model quality research focuses mainly on improving the heuristics for classical inlier counting \cite{stewart1995minpran, torr2000mlesac, torr2002bayesian}. 
LO-RANSAC \cite{chum2003locally} applies a local optimization step to promising models generated from the RANSAC sampling.
GC-RANSAC \cite{barath2018graph} extended previous local optimization and uses graph techniques to infer spatial structures and mask out outliers.
MAGSAC++ \cite{barath2019magsac, barath2020magsac++} proposes an iterative inlier counting score over a range of inlier thresholds, which reduces the sensitivity to the inlier-outlier threshold parameter. 
MQ-Net \cite{barath2022learning} combines the inlier counting score with a neural network that predicts the quality of a hypothesis from correspondence residuals.

 \input{figures/selection_plot}

\noindent
\textbf{Transformers in vision.} Since its introduction, the transformer architecture~\cite{vaswani2017attention} has become the standard in natural language processing due to its performance and simple design. Transformers are getting attention in the vision community, and have been applied successfully to image matching \cite{sun2021loftr, jiang2021cotr}, multi-view stereo \cite{he2020epipolar,ding2022transmvsnet,wang2022mvster}, or depth estimation \cite{li2021revisiting, guizilini2022multi}, among others \cite{dosovitskiy2020image, Rockwell2022}. 
Moreover, different works have been proposed to guide the cross-attention mechanism with epipolar supervision \cite{he2020epipolar, ding2022transmvsnet,wang2022mvster, guizilini2022multi},
where the most popular strategies use the epipolar attention to limit the matching search space~\cite{he2020epipolar}, or to check the consistency of a depth map \cite{guizilini2022multi}. 
\input{figures/num_corr_vs_maa}

\section{Analysis}\label{sec:motication}
In this section we explore shortcomings of the traditional RANSAC-based pose estimation approach. First, we show that correspondence-based scoring of fundamental matrix hypotheses is the major reason for RANSAC returning incorrect relative poses. Next, we show that the low number of correspondences often leads to scoring function failures motivating correspondence-free approach.
To predict \fmat quality without correspondence-based heuristics, we need a good training signal. We consider alternatives to correspondence-based scoring functions that can be used to quantify the quality of fundamental matrix hypotheses.
For our analyses, we use SuperPoint-SuperGlue (SP-SG)~\cite{detone2018superpoint, sarlin2020superglue} correspondences with MAGSAC++~\cite{barath2020magsac++}, a top-performing publicly available feature extraction, feature matching and robust estimation model. We opt for a top performing combination instead of a classical baseline, \ie, SIFT with RANSAC, to analyse a more realistic use case. We then mine image pairs with low overlapping views from validation splits of ScanNet~\cite{dai2017scannet} and MegaDepth \cite{li2018megadepth} datasets as in \cite{sarlin2020superglue, sun2021loftr} and study MAGSAC++'s behavior for 500 iterations, \ie, 500 \fmat hypotheses were generated and scored. See Section~\ref{sec:method:implementation_details} 
for more details on the validation set generation. 


\noindent
\textbf{Where do wrong solutions come from?}
In Figure~\ref{fig:selecting_F}, we show the number of times a good fundamental matrix (\fmat with pose error $<$ 10\textdegree) was selected.
We also count failures: (i) pre-scoring, \ie, no good \fmat is among hypotheses, (ii) degeneracy cases, \eg, inlier correspondences can be explained with a homography,
(iii) scoring failures, where bad \fmat was chosen by MAGSAC++ heuristic while a good \fmat was among the 500 hypotheses but had lower score. As seen in the figure, 42\% (ScanNet) and 23\% (MegaDepth) of the image pairs generate a valid \fmat but the scoring method is not able to select it. 


\noindent
\textbf{Why are wrong solutions selected?}
Inlier heuristics find fewer inliers for a good \fmat than a bad \fmat. So, what leads to a low number of inliers for a good \fmat?
Intuitively, feature matching methods
return fewer correspondences when the image pair is difficult to match, due to repetitive patterns, lack of textures, or small visual overlap, among the possible reasons.
In such scenarios, correspondences can also be highly contaminated by outliers, making it even harder to select the correct \fmat.
Figure~\ref{fig:corr_vs_maa} shows the \maa as in \cite{jin2021image} \textit{w.r.t} the number of correspondences generated by \mbox{SP-SG} and MAGSAC++.
We see the strong correlation between the number of correspondences and the accuracy of the selected model. Furthermore, the SuperGlue correspondences in this experiment are already of high-quality, as bad correspondences were filtered out. In Appendix \ref{appendix:sec:more_corr}, we show that loosening the correspondence filtering criteria, and hence, increasing the number of correspondences, does not lead to improved results. 
This observation motivates our correspondence-free \fmat hypothesis scoring approach, which leads to improvements in relative pose estimation task when correspondences are not reliable.

\input{figures/architecture_plot}


\input{tables/error_criteria_table}

\noindent
\textbf{How should we select good solutions?}
If we want to train a model that predicts fundamental matrix quality what should this model predict? As we are interested in ranking hypotheses, or potentially discard all of them if they are all wrong, the predicted value should correlate with the \emph{error in the relative pose} or the \emph{mismatch of epipolar constraint}.

The quality of relative pose is measured for rotation and translation separately. The angle of rotation between the estimated and the ground truth rotation matrices provides the rotation error. As fundamental and essential matrices are scale-invariant, the scale-invariant metric of translation error is the angular distance 
between the estimated and the ground truth translation vectors. Thus, relative pose error requires estimation of two values.

There are multiple error criteria~\cite{fathy2011fundamental} for fitting \fmats. Reprojection error (RE) measures the minimum distance correction needed to align the correspondences with the epipolar geometry. Symmetric epipolar (SED) measures geometric distance of each point to its epipolar line. Finally, Sampson (RE1) distance is a first-order approximation of RE. The benefit of these metrics is that the quality of fundamental matrices is measured with one scalar value.

So, how do we choose the best error criteria as training signal for \ours? Should it be relative pose error, or one of the fundamental matrix error criteria?
Let us assume that we have access to multiple oracle models, one oracle for each of different error criteria in Table~\ref{tab:err_criteria}. So, one oracle model that predicts the relative pose error perfectly, one oracle model that predicts SED error perfectly, \etc. Which of these oracle models provides the best \fmat scoring approach? 
To simulate evaluation of these oracle models, we use SP-SG and MAGSAC++ to mine \fmat hypotheses from validation datasets, and use the ground truth depth maps and camera poses to generate dense correspondences between image pairs to exactly compute all error criteria.
In Table~\ref{tab:err_criteria}, we show the \maa of different scoring criteria when evaluating mined fundamental matrices using oracle models in ScanNet and MegaDepth image pairs.
As can be seen in the table, all the \fmat error criteria under-perform compared to relative pose error metrics, hence \ours should be supervised by the relative rotation and translation errors. 



%
%

\section{Method}\label{sec:method}

\ours estimates the quality of a fundamental matrix hypothesis, $\mathbf{F}_i$, for the two input images, $A$ and $B$, without relying on correspondences and processing the images directly. Besides fundamental matrix quality estimation, in a calibrated setup, \ours could compute the score of the essential matrix, $\mathbf{E}_i$, by first obtaining $\mathbf{F}_i$ based on their relationship: $\mathbf{E}_i = \mathbf{K}_B^T \mathbf{F}_i \mathbf{K}_A$. Figure~\ref{fig:architecture} shows the four major components of \ours architecture: the feature extractor, the transformer, the epipolar cross-attention, and the pose error regressor. Appendix \ref{appendix:sec:architecture} provides more detailed description of our network.

\subsection{Feature Extractor} \label{sec:method:extractor}
 The feature extractor is a standard convolutional architecture as in \cite{sun2021loftr}, it follows the Unet-style network~\cite{ronneberger2015u} design with skip and residual connections~\cite{he2016deep} and computes feature maps at $1/4$ of the input resolution. Before feature extraction, we center-crop and resize the input images $A$ and $B$ to a resolution of $(H, W)$ (in our experiments we use $(256, 256)$). Images are then processed by the convolutional feature extractor to produce the C-dimensional feature maps $\mathbf{f}^A$ and $\mathbf{f}^B$.

\subsection{Transformer} \label{sec:method:transformer}
We use an $L$ multi-head attention transformer architecture following~\cite{sun2021loftr}, and alternate between self and cross-attention blocks to exploit the similarities within and across the feature maps. We denote the transformed features as $^\dagger \mathbf{f}^A$ and $^\dagger \mathbf{f}^B$.
Following the transformer nomenclature, some features are used to compute the query (Q), and potentially different features are used to compute the key (K) and the value (V). 
Q retrieves information from V based on the attention weight computed from the product of Q and K.
In the self-attention layer, the same feature map builds Q, K, and V, meanwhile, in the cross-attention layer, Q is computed from a different feature map than K and V.
We interleave the self and cross-attention block $N_t$ times.

\noindent
\textbf{Self-attention and Cross-attention layers.} To limit the computational complexity of our transformer block, we use a Linear Transformer \cite{katharopoulos2020transformers} as in \cite{sun2021loftr}. Linear Transformer reduces the computational complexity of the original Transformer from $O(N^2)$ to $O(N)$ by making use of the associativity property of matrix products and replacing the exponential similarity kernel with a linear dot-product kernel. 
Specifically, in a self-attention layer, the input feature map $\mathbf{f}'$ is used to compute Q, K and V. We concatenate the result of the attention layer with the input $\mathbf{f}'$ feature map and pass it through a two-layer MLP. The output of the MLP is then added to $\mathbf{f}'$ and passed to the next block. In the cross-attention layer, we repeat the previous process but compute Q from one feature map and K and V from the second. 

We found that positional encoding for attention layers did not improve our results, similarly to findings in \cite{li2021revisiting}, and hence, we do not use positional encodings in \ours.

\subsection{Epipolar Cross-attention} \label{sec:method:epi_attention}


Up to this point, attended feature maps, $^\dagger \mathbf{f}^A$ and $^\dagger \mathbf{f}^B$, are cached and reused. Given that \ours computes the score for every $\mathbf{F}_i$, this design assures a more practical scenario where the overhead of computing additional \fmat scores is small. 

For every \fmat hypothesis $\mathbf{F}_i$, our epipolar cross-attention mechanism embeds $\mathbf{F}_i$ together with feature maps $^\dagger \mathbf{f}^A$ and $^\dagger \mathbf{f}^B$. Every position $\mathbf{p}^A = [u, v]$ in feature map $^\dagger \mathbf{f}^A$ has a corresponding epipolar line in $^\dagger \mathbf{f}^B$ defined as $\mathbf{l}_{uv}^{A\rightarrow B} = \mathbf{F}'_i \bar{\mathbf{p}}^A$, 
where $\bar{\mathbf{p}}^A$ refers to the homogeneous coordinates of $\mathbf{p}^A$ and $\mathbf{F}'_i$ is a scaled $\mathbf{F}_i$ by a factor of $1/4$. As we consider potentially hundreds of hypotheses, the resolution of feature maps impacts run-time speed. So, we opt to define query points, $\mathbf{p}^A = [u, v]$, with a step sampling of two. This reduces even further the final feature map to a resolution of $1/8$ of the input image.

So, for every feature $^\dagger \mathbf{f}_{uv}^A \in {}^\dagger\mathbf{f}^A$ we sample $^\dagger \mathbf{f}^B$ at $D$ equidistant locations along the epipolar line $\mathbf{l}_{uv}^{A \rightarrow B}$. We start sampling where the epipolar line meets the feature map (from left to right) and use bilinear interpolation to produce $D$ features $^\dagger \mathbf{f}^B_{\mathbf{l}_{uv}}$. If sampling positions fall outside the image plane, or the epipolar line never crosses the image, we zero pad the features. Thus, we build feature volume $^\dagger \mathbf{f}^B_{i} \in [C, D, W/8, H/8]$ from feature map $^\dagger \mathbf{f}^B$ and $\mathbf{F}_i$.
We use $^\dagger \mathbf{f}^A$ to compute Q, and $^\dagger \mathbf{f}^B_i$ volume to obtain the K and V, and perform attention along the epipolar candidate points. 
Finally, we use Q, K, and V to obtain epipolar transformed features $\mathbf{f}_i^A$.
For order-invariance, we also compute  $\mathbf{f}_i^B$ by repeating these operations for $(^\dagger \mathbf{f}^B, ^\dagger \mathbf{f}^A)$ pair of feature maps and $\mathbf{F}_i^T$.

\subsection{Pose Error Regressor} \label{sec:method:regressor}
As seen in Figure \ref{fig:architecture}, the pose error regressor uses a ResNet block to extracts features from $\mathbf{f}_i^A$ and $\mathbf{f}_i^B$. Following \cite{cai2021extreme, arnold2022map}, we apply a 2D average pooling that results in two 1D vectors, $\mathbf{v}_i^{A\rightarrow B}$ and $\mathbf{v}_i^{B \rightarrow A}$,  with size $C'$. Both 1D vectors are then merged by a max pooling operator, such that different order of the input images always produce the same feature vector $\mathbf{v}_i$. An MLP layer then regresses the angular translation and rotation errors, $e_i^t$ and $e_i^R$, associated to $\mathbf{F}_i$. 

\subsection{Loss Function}
Contrary to previous binary \cite{barath2022learning} or multi-class \cite{cai2021extreme} formulation of the pose error, we experimentally found that \ours is more accurate when regressing independently the translation and rotation errors (see Section~\ref{sec:experiments:ablation}). The supervision we use is angular errors, hence, the predicted error, $e_i$, and the ground truth error, $\bar{e}_i$, are bounded between $[0^\circ, 180^\circ]$. Directly using a simple L1 loss ($l = |\bar{e}_i-e_i|$) would treat all error ranges equally. However, we are interested in accurate error estimation of all good \fmats, while not requiring being as precise when the \fmats have high pose errors. Hence, we propose to use a soft clamping of the ground truth error as well as the network prediction, such as: 
\begin{equation}
\begin{gathered}
    \mathcal{L} = |g(\bar{e}_i^t) - g(e_i^t)| + |g(\bar{e}_i^R)-g(e_i^R)|,
\label{eq:loss_function}
\end{gathered}
\end{equation}
where $g(x)$ refers to the $\tanh (x / t_s)$ function, and $t_s$ is the scaling factor that adjusts the (soft) threshold after which the accuracy of the angular errors is not important.

\input{tables/F_estimation_ScanNet}
\subsection{Implementation Details} \label{sec:method:implementation_details}
We train indoor and outdoor \ours models on ScanNet \cite{dai2017scannet} and MegaDepth \cite{li2018megadepth} datasets. To generate training and validation sets, we first extract SP-SG \cite{detone2018superpoint, sarlin2020superglue} correspondences using appropriate indoor and outdoor pre-trained models. 
We then draw minimal subsets of correspondences randomly and extract 500 two-view hypotheses for every image pair. 
For each hypothesis, we compute the angular translation ($e_t$) and rotation ($e_R$) errors using the ground truth extrinsic and intrinsic parameters. 
During training, we ensure that the ground truth hypothesis is among the 500 hypotheses. In batch generation, we cluster the hypotheses into bins based on the pose error and randomly select a bin, from which one hypothesis is uniformly sampled.

For the indoor model, we use the training splits proposed in \cite{sun2021loftr}.  
For our outdoor model, we use the MegaDepth training and validation scenes proposed in \cite{sun2021loftr}, except for scenes in the test set of the CVPR IMW 2019 PhotoTourism dataset~\cite{thomee2016yfcc100m, jin2021image} as in \cite{sarlin2020superglue}. For both datasets, we generate a total of 90,000 and 30,000 image pairs with 500 fundamental and essential matrix hypotheses for each image pair. To highlight the benefit of \ours over hard image pairs, we sample image pairs in both datasets that have a visual overlapping score between 10\% and 40\% of the image. 

We indicate the training source of \ours by (F) and (E) for fundamental and essential matrices, respectively.  
We also found that the distribution of essential matrices is very different from distribution of \fmats (see Appendix \ref{appendix:sec:dist_E_F}), so, we combine fundamental and essential matrix hypotheses 
into (F + E) training dataset.

\ours is trained end-to-end with randomly initialized weights, a learning rate of $1 \times 10^{-4}$ and a batch size of 56 image pairs. We train on four V100 GPUs and the network converges after 72 hours. For the feature extractor, we use a ResNet-18 \cite{he2016deep} as in \cite{sun2021loftr}. The transformer block uses three attention layers ($N_t=3$), and we sample $D=45$ matching candidates along the epipolar line. The loss formulation uses $t_s=25$, such that the pose errors above 40\textdegree are  (soft) clamped. At test time, \ours selects the model hypothesis that returns the minimum pose error, where the pose error for $\mathbf{F}_i$ is computed as $e_i = \max(e_i^R, e_i^t)$.

%
%

\input{sections/experiments}


%
%

\section{Conclusions}\label{sec:conclusions}
\vspace{-1pt}
We introduce \ours, a correspondence-free two-view geometry scoring method. Our experiments show that our learned model achieves SOTA results, performing exceptionally well in challenging scenarios with few or unreliable correspondences. These results suggest exciting avenues for future research. We mention some of them here.

Naive combinations of \ours with traditional inlier counting lead to even higher reliability. This suggests that more sophisticated combinations of scoring approaches could lead to further improvements. Alternatively, a better network design may improve prediction accuracy, which could supersede inlier counting in all scenarios.

The first stages of our pipeline (feature extraction, self and cross-attention blocks) form a subnetwork that precomputes feature maps for each image pair. This subnetwork architecture is similar to the latest feature extraction methods, \eg~\cite{sun2021loftr}. This means that feature map precomputation, local feature extraction, feature matching, and hypothesis generation can all be performed by this subnetwork. Training the whole network on two tasks: estimating good correspondences and two-view geometry scoring could lead to synergistic improvements on both tasks.


\begin{appendices}

Complementary to the details and experiments of the main paper, we also include additional analyses and results for \ours in this appendix. In Appendix \ref{appendix:sec:architecture}, we describe \ours blocks and layers in detail. Appendix \ref{appendix:sec:generalization} shows results of FSNet generalizing to hypotheses generated by different correspondence estimation methods. In Appendix \ref{appendix:sec:more_corr}, we prove that increasing the number of SP-SG \cite{detone2018superpoint, sarlin2020superglue} correspondences does not yield better results. Appendix \ref{appendix:sec:dist_E_F} discusses the differences in the error distributions of the generated fundamental and essential matrices of our validation set and discusses the benefits of specializing \ours for each of the tasks. In Appendix \ref{appendix:sec:sg_test_set}, we extend the experiments of the main paper and report results on the official SuperGlue \cite{sarlin2020superglue} ScanNet test set. Appendix \ref{appendix:sec:cross_entropy} gives more details about the binary cross-entropy loss formulation used in the ablation study of Table \ref{tab:loss_ablation} (Section \ref{sec:experiments:ablation}). Finally, in Appendix \ref{appendix:sec:cand_filter}, we discuss in more detail the candidate filter approach proposed in the main paper and show how to exploit and combine inlier counting heuristics, \ie, MAGSAC++ \cite{barath2020magsac++}, with \ours.

\section{\ours Architecture}
\label{appendix:sec:architecture}
This section extends on the implementation details of \ours architecture from the main paper.
Table \ref{tab:extractor} introduces the layers within the feature extractor block that we use for computing the features $\mathbf{f}^A$ and $\mathbf{f}^B$ from images $A$ and $B$. Input images have a resolution of $256 \times 256$, and the feature extractor block outputs feature maps of size $128 \times 64 \times 64$. As seen in Table \ref{tab:extractor}, the feature extractor is composed of ResNet-18 \cite{he2016deep} blocks, where every block is based on $3\times3$ convolutions, batch normalization layers \cite{ioffe2015batch}, ReLU activations \cite{agarap2018deep}, and a residual connection. After the ResNet blocks, we upsample the feature maps twice and create skip connections with previous layers following a UNet \cite{ronneberger2015u} architecture design. Please refer to Table~\ref{tab:extractor} to see which layers are combined by the skip connections. A final convolution layer with a batch normalization layer and a Leaky-ReLU activation \cite{xu2015empirical} generates the feature maps $\mathbf{f}^A$ and $\mathbf{f}^B$.

Once feature maps are extracted, we feed them to our transformer architecture (see Section \ref{sec:method:transformer}). The transformer computes the transformed features $^\dagger \mathbf{f}^A$ and $^\dagger \mathbf{f}^B$, which exploit the self and cross-similarities across the feature maps. For our transformer architecture, we follow the design of the Linear Transformer \cite{katharopoulos2020transformers, sun2021loftr}. We use three attention layers ($N_t=3$), where every self and cross-attention layer has eight attention heads. The transformer outputs $^\dagger \mathbf{f}^A$ and $^\dagger \mathbf{f}^B$, which are then stored and reused for every input $F_i$ hypothesis. 
\input{tables/supp_table_architecture}

We embed the two-view geometry into the features through an epipolar cross-attention block. The epipolar cross-attention takes $^\dagger \mathbf{f}^A$, $^\dagger \mathbf{f}^B$, and $F_i$ to guide the attention between the two feature maps. The epipolar cross-attention layer applies cross-attention along the epipolar line. For every query point, we sample $D=45$ positions along its corresponding epipolar line 
, and hence, attention is done only to the $D$ sampled positions. 
Some sampling positions might be outside of the feature plane, \textit{e.g.}, epipolar line never crosses the feature map. Thus, in those cases, we pad the positions with zeros, such that they do not contribute when computing the attended features. In the transformer Softmax, those positions will not matter as their contribution to the soft-aggregation is zero.
To reduce the feature map size, which contributes towards faster processing time, we query points every two positions. The epipolar cross-attention layer outputs $\mathbf{f}_i^A$ and $\mathbf{f}_i^B$ at 1/8 of the input image resolution ($128 \times 32 \times 32$). 

Table \ref{tab:arch_err_regressor} shows the details of the pose error regressor block. The pose error regressor takes the epipolar attended features ($^\dagger \mathbf{f}^A$, $^\dagger \mathbf{f}^B$) and predicts the translation and rotation errors ($e_i^t$ and $e_i^R$) associated with $F_i$. Similar to the feature extractor, the pose error regressor uses ResNet-18 blocks to process the features. After processing the features, a 2D average pooling is applied to create $\mathbf{v}_i^{A\rightarrow B}$ and $\mathbf{v}_i^{B \rightarrow A}$. To enforce image order-invariance, we merge the two vectors with a Max-Pooling operator. A final MLP uses the output of the Max-Pool $v_i$ to predict the pose errors associated with $F_i$ hypothesis.

\section{\ours with LoFTR and SIFT correspondences}
\label{appendix:sec:generalization}
\input{tables/loftr_table}
\ours is trained using hypotheses generated with \mbox{SP-SG}~\cite{detone2018superpoint, sarlin2020superglue} correspondences. While \ours does not rely on correspondences to do scoring, the hypothesis pool is generated from correspondences. To show the generalization capability of our network, 
we extend the previous experiments and show results of \ours generalizing to hypotheses generated by different correspondence estimation methods. We choose SIFT~\cite{lowe1999object} and LoFTR~\cite{sun2021loftr}. We use Kornia \cite{riba2020kornia} library to compute LoFTR correspondences, and SIFT matches are filtered with the mutual nearest neighbor check and Lowe's ratio test~\cite{lowe1999object}.\\

Table~\ref{tab:loftr_results} shows the results of MAGSAC++ and \ours combined with SIFT and LoFTR. As a reference, refer to Table \ref{tab:F_estimation_table} for \mbox{SP-SG} results.
We observe that (i) LoFTR performance is lower than \mbox{SP-SG}, and (ii) MAGSAC++ alone or in combination is struggling, leaving \ours (alone) as the winner. We believe that one possible cause for LoFTR's lower performance is the distribution of our test set, which uses image pairs with very low image overlap (10\%-40\%), and hence, it is different from the image pairs used for LoFTR training.
Besides LoFTR results, we also show that \ours can be paired with SIFT. 
Although the distribution of hypotheses generated by SIFT is potentially different, \ours ranks them successfully achieving similar mAA scores as MAGSAC++, while reducing the median pose error. 

\section{More Correspondences for Difficult Image Pairs}
\input{tables/SP_more_corr}
\input{figures/supp_F_E_dist}
\label{appendix:sec:more_corr}
In Section \ref{sec:motication}, we mention that loosening the filtering criteria of SuperGlue, and thus increasing the number of correspondences provided to MAGSAC++, does not lead to improvements in scores.

Indeed, SuperGlue filters correspondences by considering the matching confidence of the correspondences, where the threshold is $0.2$. However, given that the number of correspondences impacts the performance of correspondence-based scoring methods (Figure \ref{fig:corr_vs_maa}), would more correspondences improve the estimation of the fundamental or essential matrices?
To investigate this, we varied the filtering threshold of SuperGlue to increase the number of correspondences that go into the RANSAC loop when needed.
So, we compile a list of image pairs that initially have $\leq 100$ correspondences. These are image pairs in our ScanNet test set that are used to report scores of \mbox{0-100} splits in the tables \ref{tab:F_estimation_table}, \ref{tab:outdoor_tablev2}, and \ref{tab:loss_ablation} from the main paper. Then, we lower the threshold progressively by steps of $0.04$ until either; we obtain more than 100 correspondences, or we reduce the threshold to $0.0$ (hence, we will use all possible matches). We refer to this SuperGlue as SP-SG*, and show in Table \ref{tab:SP_more_corr} its evaluation with MAGSAC++. As mentioned previously, an increased number of correspondences does not consistently improve scores for inlier counting baselines. Indeed, MAA scores are almost the same for the fundamental matrix estimation task and slightly worse for essential matrix estimation task. The median errors are lower for the fundamental matrix estimation when more correspondences are used, but the errors increase for essential matrix estimation. Note that the numerical results have elements of randomness to them, as a different number of correspondences produces different pools of 500 random hypotheses.

\section{Distribution of F and E Hypotheses}
\label{appendix:sec:dist_E_F}
\input{tables/SuppMat_F_E_ScanNet}

When evaluating \ours in the fundamental or essential matrix estimation task, we observe that specializing \ours for a specific task was more effective  than training the architecture to solve both tasks at the same time (Table \ref{tab:F_estimation_table}). This behaviour is explained by looking into Figure \ref{fig:supp_F_vs_E_dist}. The figure shows the error distributions of the generated fundamental and essential matrices of our validation set. For completeness, we report the translation and rotation error separately, as well as the distributions for ScanNet and MegaDepth datasets. 

We observe that the distributions of the errors are different when estimating fundamental or essential matrices, where fundamental matrices tend to have a wider range of pose errors, thus, making it more difficult to select accurate estimates. This observation is also in line with the results of Table \ref{tab:F_estimation_table}, where we show that \ours was more effective when trained for a specific task, instead of using matrices from both distributions (F + E).

\section{SuperGlue Test Set}
\label{appendix:sec:sg_test_set}
Results in the main paper are reported on our own test set for Scannet.  We generate this test set split such that image pairs have a distribution of image overlaps that focuses on hard-to-handle image pairs (please see blue bars in Figure~\ref{fig:overlap_dist_test}). However, SuperGlue also published image pairs that they used for evaluation. For completeness and easy comparison, we provide evaluations of \ours on SuperGlue's test set. 

\input{figures/overlap_dist_tests}

We benchmark our performance on the standard SuperGlue test set~\cite{sarlin2020superglue}, with results shown in Table~\ref{tab:sup_F_E_estimation_table}. \ours can be seen as helpful there, but that table's scores are dominated by ``easy pairs'': we see that the regular SuperGlue test set has many image pairs with high overlap scores. That SuperGlue test set has limited exposure to scoring-failure cases, as can be seen from the overlapping statistics of the SuperGlue test set in Figure~\ref{fig:overlap_dist_test}. 

We observe the same trend in Table~\ref{tab:F_estimation_table} and Table \ref{tab:sup_F_E_estimation_table}. \ours improves over MAGSAC++ in the \mbox{0-100} split, both for fundamental and essential matrix estimation tasks. Similarly, the combined approaches with \ours provide the best scores.

\section{\ours Trained with Cross Entropy Loss}
\input{figures/candidate_filter}
\input{tables/unimodal_vs_multimodal}
\input{figures/uni_vs_multi}
\label{appendix:sec:cross_entropy}
Experiments in Section \ref{sec:experiments:ablation} (Table \ref{tab:loss_ablation}) show \ours trained with the binary cross-entropy loss. We treat the hypotheses ranking problem as a classification problem. Hypotheses with pose error $<10^\circ$ were labeled as ``correct'' and others as ``incorrect'', so
\begin{equation}
y_i = 
    \begin{cases}
    1,              & \text{if } \text{max}(e_i^R, e_i^t) < 10^\circ\\
    0,              & \text{otherwise.}
\end{cases}
\end{equation}

This means that out of 500 hypotheses generated for an image pair, multiple hypotheses can be labeled as ``correct''. As we are interested in ranking hypotheses during scoring, we can use the network's confidence in the predicted binary decision to provide ranking among ``correct'' hypotheses. Furthermore, we are not interested in the relative ranking of ``incorrect'' hypotheses. To reflect these nuances, we incorporate the network's predicted confidence in the loss function. Inspired by Barath~\etal's~\cite{barath2022learning} loss function, we modify the cross entropy loss to be
\begin{align}
    \mathcal{L} = & - (1 + f(\mathbf{F}_i)) ^ w \nonumber \\
    &\left[
        y_i \log f(\mathbf{F}_i) + (1 - y_i) \log(1 - f(\mathbf{F}_i))
    \right],
\end{align}
where $f(\mathbf{F}_i)$ is \ours network's confidence of $\mathbf{F}_i$ being a correct hypothesis, and the $w=2$ is the weight of the networks' confidence in the loss function. Low values of $f(\mathbf{F}_i)$ means that the weighting coefficient $(1 + f(\mathbf{F}_i)) ^ w$ is low, while high confidence values of $f(\mathbf{F}_i)$ provide higher weighting to the cross entropy loss.

\section{Filtering hypotheses with MAGSAC++}
\label{appendix:sec:cand_filter}
As mentioned in the main paper, we can discard not promising hypotheses by looking at their MAGSAC++ scores. This filtering approach exploits the useful information that inlier counting brings to well defined set of correspondences, while also removes easy to detect outliers with such heuristics. Besides cleaning the pool of hypotheses, it also provides a speedup opportunity since \ours only needs to be run for a subset of fundamental/essential matrices.

In Table \ref{tab:uni_vs_multi}, we analyze the top scoring hypotheses returned by MAGSAC++. Specifically, we look into the top 5 MAGSAC++ models, and analyze whether they present a multimodal or unimodal distribution. We define the distribution as unimodal if the minimum and maximum distance between the hypotheses (within the top 5) is below 10\textdegree. In the unimodal scenario, we further look if most accurate hypothesis was returned either by MAGSAC++ or \ours. Similarly, in the multimodal scenario (there are hypotheses with more than 10\textdegree difference), we identify whether 1) both methods return a valid hypothesis (with a pose error below 10$^{\circ}$ \textit{w.r.t.} ground-truth pose), 2) an incorrect hypothesis ($e>$ 10\textdegree) was selected by both methods, 3) only \ours or 4) MAGSAC++ selected a correct hypothesis. We observe that \ours returns more accurate hypotheses than MAGSAC++, in both unimodal and multimodal distribution scenarios. 
Moreover, in Figure \ref{fig:candidate_filter}, we show the impact of varying the number of hypotheses to score by \ours. On the left side, when only using 1 hypothesis to score, results correspond to MAGSAC++ method, meanwhile, on the right end, when using all 500 possible hypotheses, results indicate original \ours results. We see in the ScanNet and MegaDepth datasets that even refining the score of a few hypotheses brings improvements to MAGSAC++. Moreover, in Figure \ref{fig:uni_vs_multi}, we show examples of different top scoring hypotheses returned by MAGSAC++, and indicate if they belong to an unimodal or multimodal distribution.

{\small
\bibliographystyle{ieee_fullname}
\bibliography{egbib}
}

\end{appendices}

\end{document}

%% file: figures/teaser_fig.tex
\begin{figure}[t]
  \centering
   \includegraphics[width=1.0\linewidth]{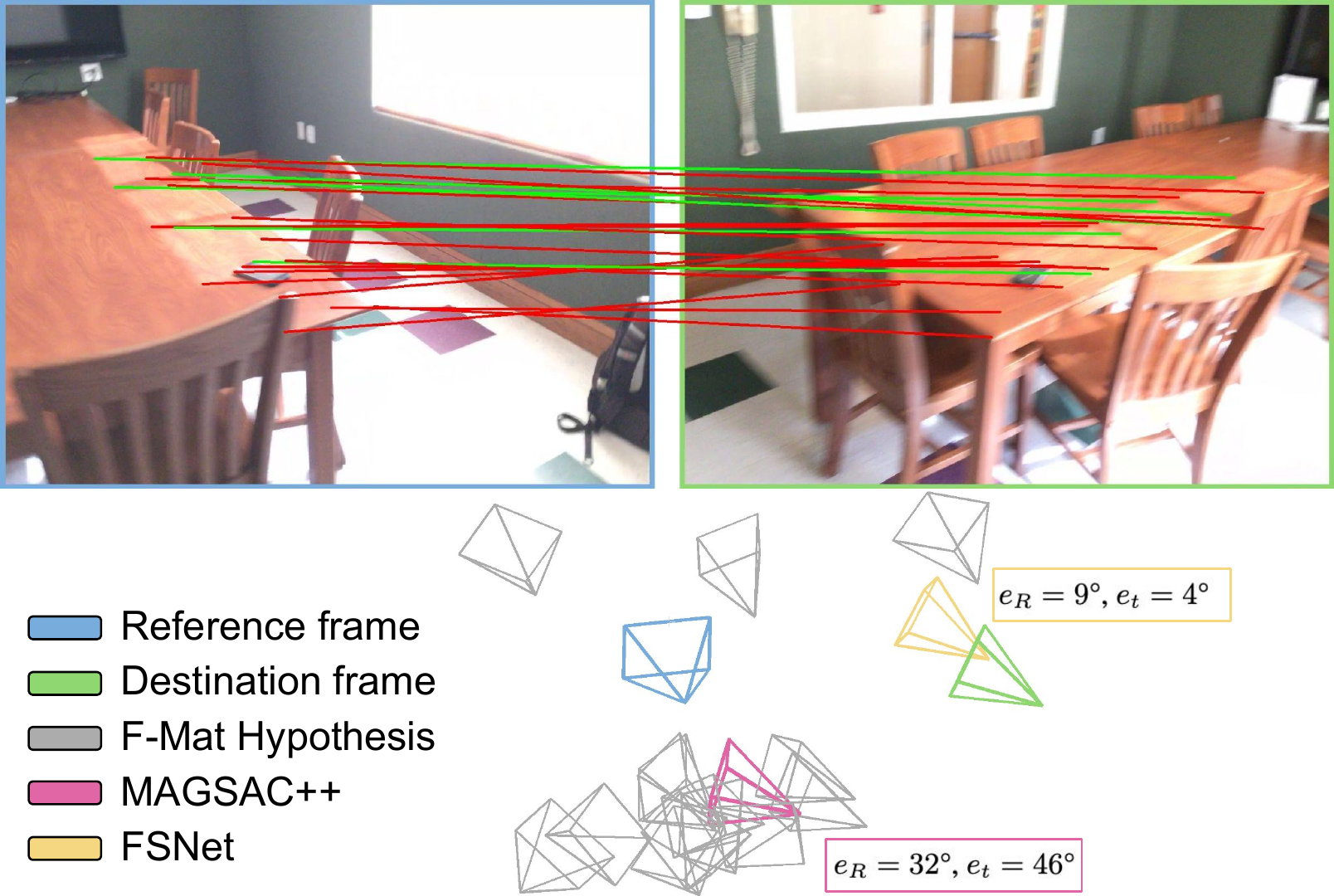}

   \caption{Example where SuperPoint-SuperGlue~\cite{detone2018superpoint, sarlin2020superglue} correspondences are highly populated by outliers, but there are still enough inliers to produce a valid fundamental matrix hypothesis. In such scenarios with unreliable correspondences, current top scoring methods fail (MAGSAC++~\cite{barath2020magsac++}), while our proposed \ours model, a correspondence-free scoring approach, is able to pick out the best \fmat.  
   }
   \label{fig:teaser}
\end{figure}



%% file: figures/selection_plot.tex
\begin{figure}[t]
  \centering
   \includegraphics[width=1.0\linewidth]{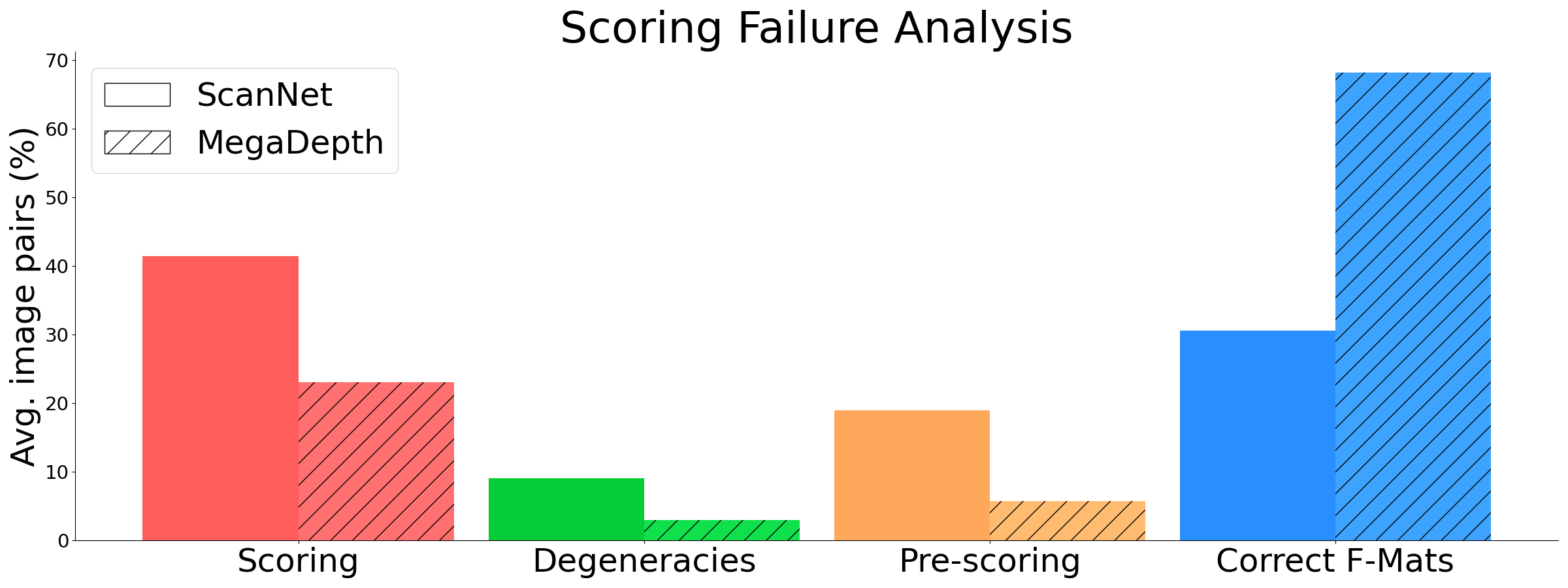}
   \caption{\textbf{Failure cases analysis}. We plot for ScanNet and MegaDepth datasets the percentage of image pairs for which a good \fmat (\fmat with pose error below 10\textdegree) was not selected due to scoring failure, degenerate case, or pre-scoring failure, \textit{i.e.}, there was not a valid \fmat in the hypothesis pool. We see that the failures center on the scoring function, it is  especially difficult to score hypotheses in the indoor scenario (ScanNet), where local features suffer more \cite{sun2021loftr}.} 
   \label{fig:selecting_F}
\end{figure}

%% file: figures/num_corr_vs_maa.tex
\begin{figure}[t]
  \centering
   \includegraphics[width=1.0\linewidth]{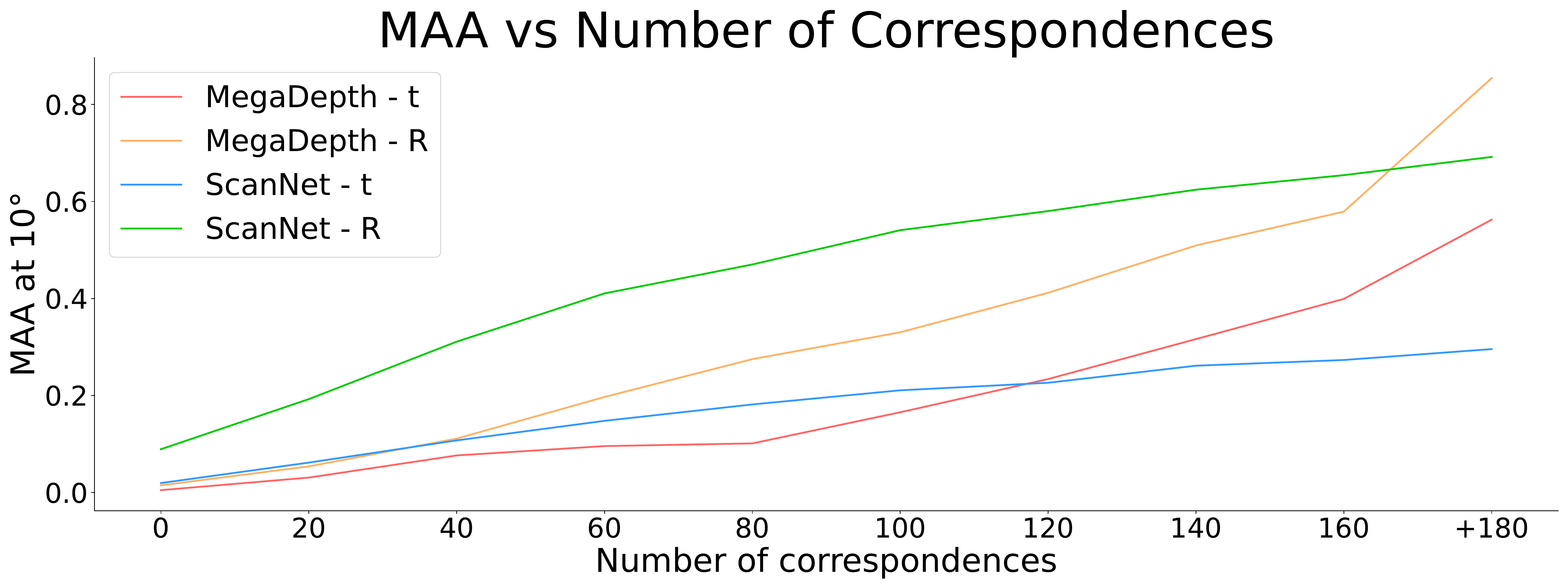}

   \caption{\textbf{\maa vs number of correspondences}. We show that the number of correspondences has a high impact on the precision of the fundamental matrix. The latest matchers already filter out the correspondence outliers \cite{sarlin2020superglue}, and naturally, the number of correspondences correlates with the difficulty of correctly matching two images. Thus, we see that the number of correspondences is a good indicator to decide when to trust the correspondences.}
   \label{fig:corr_vs_maa}
\end{figure}

%% file: figures/architecture_plot.tex
\begin{figure*}[t]
  \centering
   \includegraphics[width=0.95\linewidth]{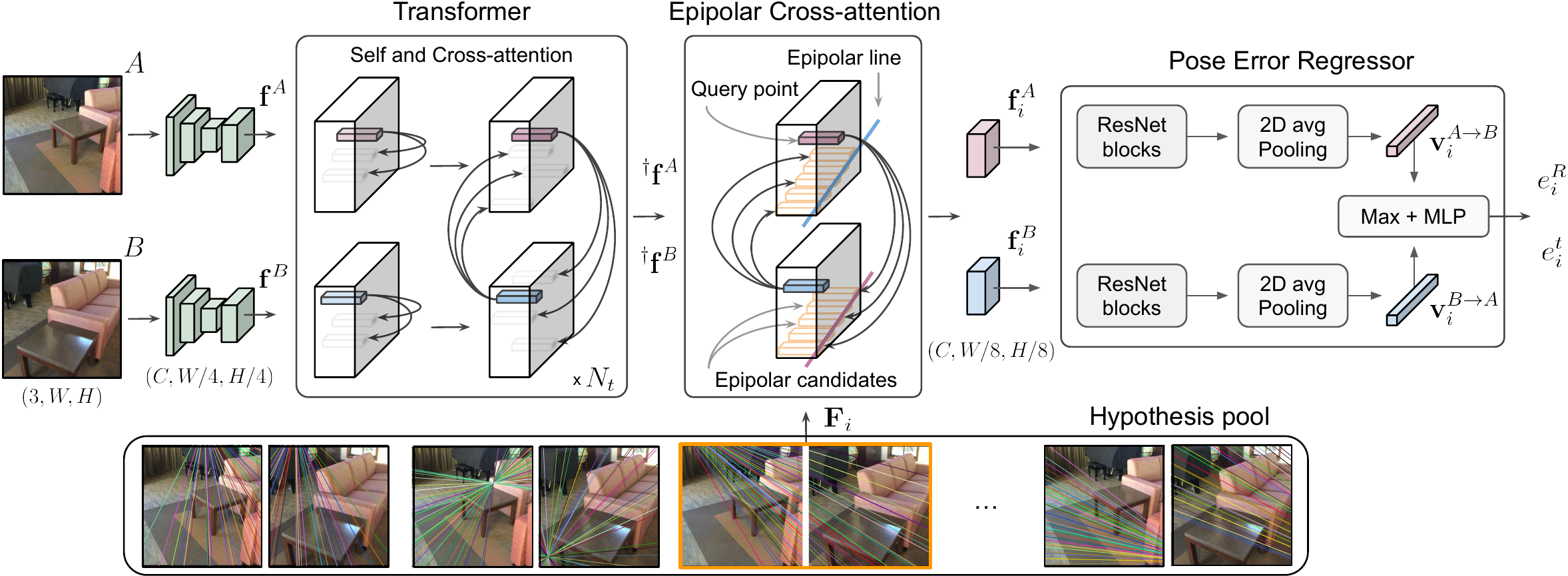}

   \caption{\textbf{\ours architecture} has four components. 1. A CNN feature extractor computes the feature maps $\mathbf{f}^A$ and $\mathbf{f}^B$ from input images $A$ and $B$ at 1/4 of input resolution (Section \ref{sec:method:extractor}). 2. The extracted features are then processed by the transformer block, which contains $N_t$ self and cross-attention layers. The transformer outputs $^\dagger \mathbf{f}^A$ and $^\dagger \mathbf{f}^B$, which are stored and reused for every $\mathbf{F}_i$ hypothesis (Section \ref{sec:method:transformer}). 3. The epipolar cross-attention layer applies cross-attention along the epipolar lines, and embeds $\mathbf{F}_i$ into the feature maps $\mathbf{f}_i^A$ and $\mathbf{f}_i^B$. Epipolar attention is done every two positions, reducing the final feature maps to an 1/8 of the input resolution (Section \ref{sec:method:epi_attention}). 4. The pose regressor applies a ResNet and a 2D average pooling block, and outputs $\mathbf{v}_i^{A\rightarrow B}$ and $\mathbf{v}_i^{B \rightarrow A}$. The vectors are combined through a max pooling operator, and the final MLP layer predicts the $e^R_i$ and $e^t_i$ errors associated to $\mathbf{F}_i$ (Section \ref{sec:method:regressor}).} 
   \label{fig:architecture}
\end{figure*}



%% file: tables/error_criteria_table.tex
\begin{table}[t]
\footnotesize
\begin{center}
\begin{tabular}{c c c l c c}
\multicolumn{1}{c}{} & \multicolumn{5}{c}{\textbf{mAA at 10\textdegree}}\\ 
\cline{2-6}
\noalign{\smallskip}
\multicolumn{1}{c}{} & \multicolumn{2}{c}{\textbf{ScanNet}} & \multicolumn{1}{c}{} & \multicolumn{2}{c}{\textbf{MegaDepth}}\\ 
\cline{2-3}
\cline{5-6}
\noalign{\smallskip}
 & R & t && R & t\\
\hline \noalign{\smallskip}
GT Pose error & \textbf{0.75} & \textbf{0.72} && \textbf{0.94} & \textbf{0.89} \\
SED & 0.57 & \underline{0.48} && 0.86 & \underline{0.78} \\
RE1 & \underline{0.74} & 0.41 && \underline{0.91} & \underline{0.78} \\
Epi. Distance & 0.57 & 0.40 && 0.89 & 0.76 \\
MAGSAC++ & 0.51 & 0.22 && 0.80 & 0.55 \\
\end{tabular}
\end{center}
\normalsize
\vspace{-1em}
\caption{\textbf{Error criteria evaluation.}
Fundamental matrices are generated with SP-SG \cite{detone2018superpoint, sarlin2020superglue} and MAGSAC++ \cite{barath2020magsac++}, and evaluated under densely projected correspondences with ground-truth camera poses and depth maps on our validation set. GT Pose error uses directly the error associated to the fundamental matrix to rank them, being the upper-bound of the scoring function.}
\label{tab:err_criteria}
\end{table}


%% file: tables/F_estimation_ScanNet.tex
\begin{table*}[ht]
\footnotesize
\begin{center}
\hspace{-1em}
\begin{tabular}{c c l c  l c  l c l c  l  c}
\multicolumn{1}{c}{} & \multicolumn{3}{c}{\textbf{0-100}} & \multicolumn{1}{c}{} & \multicolumn{3}{c}{\bf{100-Inf}} & \multicolumn{1}{c}{} & \multicolumn{3}{c}{\bf{All}}\\ 
\cline{2-4}
\cline{6-8}
\cline{10-12}
\noalign{\smallskip}
\multicolumn{1}{c}{} & \multicolumn{1}{c}{\textbf{\maa} $\uparrow$} & \multicolumn{1}{c}{} & \multicolumn{1}{c}{\textbf{\mediandeg} $\downarrow$} & \multicolumn{1}{c}{} & \multicolumn{1}{c}{\textbf{\maa} $\uparrow$} & \multicolumn{1}{c}{} & \multicolumn{1}{c}{\textbf{\mediandeg} $\downarrow$} & \multicolumn{1}{c}{} & \multicolumn{1}{c}{\textbf{\maa} $\uparrow$} & \multicolumn{1}{c}{} & \multicolumn{1}{c}{\textbf{\mediandeg} $\downarrow$}\\ 
\cline{2-2}
\cline{4-4}
\cline{6-6}
\cline{8-8}
\cline{10-10}
\cline{12-12}
\noalign{\smallskip}
 & R / t / max(R, t) && $e_R$ / $e_t$ && R / t / max(R, t)  && $e_R$ / $e_t$ && R / t / max(R, t)  && $e_R$ / $e_t$\\
\hline \noalign{\smallskip}
Fundamental & &&  &&  &&  &&  && \\
\cline{1-1}
\noalign{\smallskip}
RANSAC \cite{fischler1981random} & 0.15 / 0.07 / 0.04 && 17.97 / 30.28 && 0.24 / 0.06 / 0.05 && 11.91 / 37.03 && 0.17 / 0.07 / 0.04 && 15.54 / 32.18 \\
MAGSAC++ \cite{barath2020magsac++} & 0.28 / 0.12 / 0.08 && 9.19 / 24.38 && \underline{0.63} / \underline{0.29} / \underline{0.27} && \underline{2.80} / \underline{8.66} && 0.38 / 0.17 / 0.14 && 6.35 / 17.89 \\
\ours (F) & \underline{0.29} / \textbf{0.19} / \underline{0.12} && \underline{7.98} / \textbf{13.30} && 0.52 / 0.25 / 0.21 && 4.16 / 9.62 && 0.36 / 0.21 / 0.15 && 6.52 / 12.05 \\
\ours (F + E) & \underline{0.29} / \underline{0.18} / 0.11 && 8.01 / \underline{14.05} && 0.50 / 0.24 / 0.20 && 4.35 / 10.26 && 0.35 / 0.20 / 0.14 && 6.64 / 12.78 \\
\hdashline\noalign{\smallskip}
w/ Corresp. filter & \underline{0.29} / \textbf{0.19} / \underline{0.12} && \underline{7.98} / \textbf{13.30} && \underline{0.63} / \underline{0.29} / \underline{0.27} && \underline{2.80} / \underline{8.66} && \underline{0.39} / \underline{0.22} / \underline{0.16} && \underline{6.09} / \underline{11.59} \\
w/ Candidate filter & \textbf{0.33} / \underline{0.18} / \textbf{0.13} && \textbf{7.48} / 14.91 && \textbf{0.66} / \textbf{0.36} / \textbf{0.32} && \textbf{2.69} / \textbf{6.50} && \textbf{0.43} / \textbf{0.23} / \textbf{0.19} && \textbf{5.38} / \textbf{11.39} \\
\hline
\hline \noalign{\smallskip}

Essential & &&  &&  &&  &&  && \\
\cline{1-1}
\noalign{\smallskip}

EssNet \cite{zhou2020learn}  & - && - && - && - && 0.01 / 0.02 / 0.01 && 48.64 / 52.95 \\
Map-free \cite{arnold2022map}  & - && - && - && - && 0.39 / 0.13 / 0.09 && 5.75 / 14.20 \\
RANSAC \cite{fischler1981random} & 0.27 / 0.16 / 0.13 && 9.78 / 16.64 && 0.63 / 0.38 / 0.33 && 3.03 / 6.15 && 0.37 / 0.24 / 0.19 && 6.55 / 11.68 \\
MAGSAC++ \cite{barath2020magsac++} & 0.29 / 0.19 / 0.14 && 8.90 / 16.00 && \underline{0.65} / \underline{0.41} / \underline{0.38} && \underline{2.57} / \underline{5.65} && 0.40 / 0.26 / 0.21 && 5.95 / 10.96 \\
\ours (E) & \textbf{0.36} / \textbf{0.25} / \textbf{0.18} && \textbf{6.34} / \textbf{10.51} && 0.61 / 0.35 / 0.31 && 3.22 / 6.78 && \textbf{0.44} / \underline{0.28} / \underline{0.22} && \underline{5.13} / \underline{8.95} \\
\ours (F + E) & \underline{0.35} / \underline{0.24} / \underline{0.17} && \underline{6.70} / \underline{10.65} && 0.60 / 0.33 / 0.30 &&3.41 / 7.15&& \underline{0.43} / 0.27 / 0.21 && 5.26 / 9.21\\
\hdashline\noalign{\smallskip}
w/ Corresp. filter &\textbf{0.36} / \textbf{0.25} / \textbf{0.18} && \textbf{6.34} / \textbf{10.51} && \underline{0.65} / \underline{0.41} / \underline{0.38} && \underline{2.57} / \underline{5.65} && \textbf{0.44} / \textbf{0.29} / \textbf{0.23} && \textbf{5.07} / \textbf{8.75}\\
w/ Candidate filter & 0.33 / 0.22 / 0.16 && 7.71 / 13.41 && \textbf{0.69} / \textbf{0.44} / \textbf{0.40} && \textbf{2.40} / \textbf{5.26} && \textbf{0.44} / \underline{0.28} / \textbf{0.23} && 5.14 / 9.48 \\

\end{tabular}
\end{center}
\normalsize
\vspace{-0.8em}
\caption{\textbf{Fundamental and essential matrix estimation on ScanNet}. We compute the \maa and Median error (\textdegree) metrics in the test split of ScanNet, and divide the pairs of images based on the number of SP-SG correspondences to highlight the benefits of \ours, which results in 3,522 (0-100) and 1,478 (100-Inf) image pairs. In fundamental and essential estimation, we see that when number of correspondences is small (0-100), \ours provides a more robust solution than competitors. In the overall split (All), \ours obtains more precise rotation errors for the fundamental estimation tasks, while outperforming in all metrics in essential estimation. Moreover, we also show how \ours and MAGSAC++ can be easily combined to obtain more reliable approaches. 
}
\label{tab:F_estimation_table}
\end{table*}

%% file: sections/experiments.tex
\section{Experiments}\label{sec:experiments}
This section presents results for different fundamental and essential hypothesis scoring methods. We first compute correspondences with SP-SG \cite{detone2018superpoint, sarlin2020superglue}, and then use the universal framework USAC \cite{raguram2012usac} to generate hypotheses with a uniformly random minimal sampling. Besides the inlier counting of RANSAC \cite{fischler1981random}, we also use the state-of-the-art MAGSAC++ scoring function \cite{barath2020magsac++}, and compare them both to \ours scoring method. To control for randomness, we use the same set of hypotheses generated for each image pair for the evaluation of all methods. 
We refine all the selected hypotheses by applying LSQ fitting followed by the Levenberg-Marquardt \cite{more1978levenberg} optimization to the correspondences that agree with the best hypothesis model. 

To evaluate the scoring methods, we decompose the hypothesis (fundamental or essential matrix) into rotation matrix and translation vectors, and compute the angular errors ( $e_R$ and $e_t$) w.r.t ground truth rotation and translation. We report the mean Average Accuracy (mAA), which thresholds, accumulates, and integrates the errors up to a maximum threshold (in our experiments 10\textdegree) \cite{jin2021image}. The integration over the accumulated values gives more weight to accurate poses and discards any camera pose with a relative angular error above the maximum threshold. Besides the mAA, we report the median errors of the selected models. 

\subsection{Indoor Pose Estimation}
Evaluations on the Scannet test set are shown in Table \ref{tab:F_estimation_table}. Motivated by the analysis in Section~\ref{sec:motication}, and to emphasize the benefits of \ours, we also report results on two types of image pairs: image pairs with up to 100 input correspondences and image pairs with more than 100 input correspondences. The threshold is chosen based on our validation, where 64\% of the image pairs do not have more than 100 correspondences, and hence, top-performing methods can struggle.

\input{figures/merge_pred_plot}
We see that in the few-correspondences split (0-100), all \ours models outperform RANSAC and MAGSAC++, being especially effective on the angular translation error metrics.
Overall, \ours returns the best mAA score and median error for translation, while having comparable results to MAGSAC++ on rotation, both for fundamental and essential matrix evaluations. 
We also observe that \ours (F+E) does not score as well as models trained on F or E hypotheses only, further indicating differences in the distribution of fundamental and essential matrices.

For essential matrix evaluations, we also report scores of relative pose regression methods (RPR), \ie, EssNet \cite{zhou2020learn} and Map-free \cite{arnold2022map}, which also do not rely on correspondences. \ours (E) obtains the best mAA scores and median errors in the full test set (All). 
RPR methods do not compute as accurate poses as RANSAC-based methods or \ours. 

As MAGSAC++ is very effective when SP-SG is able to extract sufficient number of correspondences (100-Inf), we propose two approaches to combine both methods. In \textit{Corresp. filter}, we use the number of correspondences such that if the number is below 100, we use \ours to score, otherwise, we use MAGSAC++.
In \textit{Candidate filter}, we first select top hypotheses using MAGSAC++, and then 
 we use \ours to choose the best hypothesis out of the selected hypotheses. 
We use \ours to score the top 10 hypotheses for fundamental, and top 20 hypotheses for essential based on a hyperparameter search on the validation set. 
Moreover, in Figure~\ref{fig:fused_preds}, we show the percentage of correct \fmats, \ie, 
hypotheses with a pose error under given thresholds (5\textdegree, 10\textdegree, and 20\textdegree), and see how combining both methods always maximizes the number of correct \fmats.
Moreover, we observe that in the more flexible regime (20\textdegree) the \ours performance is almost identical to the combinations of both methods, indicating that although \ours estimations are not as accurate as MAGSAC++, there are fewer catastrophic predictions. We link this behaviour with the nature of our method, which uses low resolution feature maps to predict the pose errors, and hence, lacks precision. On the other hand, MAGSAC++ relies on potentially ill-conditioned sets of correspondences,
resulting in occasional catastrophic poses.


\subsection{Outdoor Pose Estimation}
\input{tables/phototourism_table}
In Table \ref{tab:outdoor_tablev2}, we show the mAA and median errors of the fundamental and essential matrix estimation tasks on the PhotoTourism test set. Analogous to the indoor evaluation, we split the image pairs based on the number of SP-SG correspondences. Contrary to the indoor scenario, there are only 1.5\% of the image pairs in the test set that have fewer than 100 SP-SG correspondences. This is in line with the analysis of Figure \ref{fig:selecting_F}, where scoring failures are not as common in outdoor dataset as in indoor dataset. There are multiple explanations of these phenomena: (i) walls and floors are mostly untextured surfaces which lead to fewer reliable correspondences for indoor images, (ii) indoor surfaces are closer to the camera, so small camera motion can result in large parts of the scene not being visible from both cameras, which could result in fewer correspondences, (iii) outdoor datasets are generated with SfM methods that rely on 2D-2D correspondences estimated with RANSAC, while indoor datasets rely on RGBD SLAM with dense geometry alignment, so inherent algorithmic bias could result in fewer difficult image pairs in outdoor datasets~\cite{brachmann2021limits}.

In the fundamental estimation task, \ours scoring approach obtains the most accurate poses in the low number of correspondences split (0-100), while still outperforming RANSAC inlier counting approach in the full test set. In the essential estimation task, even though we observe that correspondence-based scoring approaches, \ie, RANSAC or MAGSAC++, outperform \ours, 
they still benefit from combining their predictions with \ours.

\subsection{Understanding \ours} \label{sec:experiments:ablation}

\input{tables/ablation_loss}
\noindent
\textbf{Loss supervision ablation.}
In Table \ref{tab:loss_ablation}, we experimentally verify our analysis of supervision signals in Section~\ref{sec:motication}. We show that \ours trained with Eq.~\ref{eq:loss_function} outperforms \ours trained using RE1 supervision (best epipolar constraint error criteria among oracle models in Table~\ref{tab:err_criteria}) and \ours trained using $L1$ norm without soft clamping. We also train \ours using binary cross-entropy (CE), where fundamental matrix hypothesis $\mathbf{F}_i$ is deemed correct if $e_i<10^\circ$ and incorrect otherwise. We follow~\cite{barath2022learning} and incorporate the network uncertainty in the cross-entropy loss; please see Appendix \ref{appendix:sec:cross_entropy} for details. 


\noindent
\textbf{Failures}. Figure \ref{fig:failures} shows failure cases for \ours hypothesis selection. 
As \ours needs to run on hundreds of hypotheses in a reasonable time, the network is designed to work on low resolution images. Unfortunately, at low resolution, matching features can seem degenerate. Looking into the failures, we observe that the \fmats selected by \ours have a valid epipolar geometry, \ie, epipolar lines in image $B$ cross the surrounding regions of corresponding points from image $A$, 
and hence, the low resolution features ($(W/8, H/8) = (32, 32)$) given to the pose error regressor lack the precision to produce a good estimate.

\noindent
\textbf{Time analysis.} Given a pair of images and \mbox{500 hypotheses}, \ours scores and ranks them in 0.97s on a machine with one V100 GPU. Running the feature extraction and transformer subnetworks takes 138.28ms. This is precomputed once and reused for every hypothesis. The hypotheses are queried in batches, so the epipolar cross-attention and pose error regressor subnetworks run in 167.13ms for a batch of 100 fundamental matrices.
Besides, strategies such as 
\textit{Candidate filter} can reduce \ours time by only scoring promising hypotheses, \textit{e.g.}, for fundamental estimation it only adds 150ms on top of MAGSAC++ runtime (71ms).
\input{figures/failures_plot}

%% file: figures/merge_pred_plot.tex
\begin{figure}[t]
  \centering
   \includegraphics[width=1.0\linewidth]{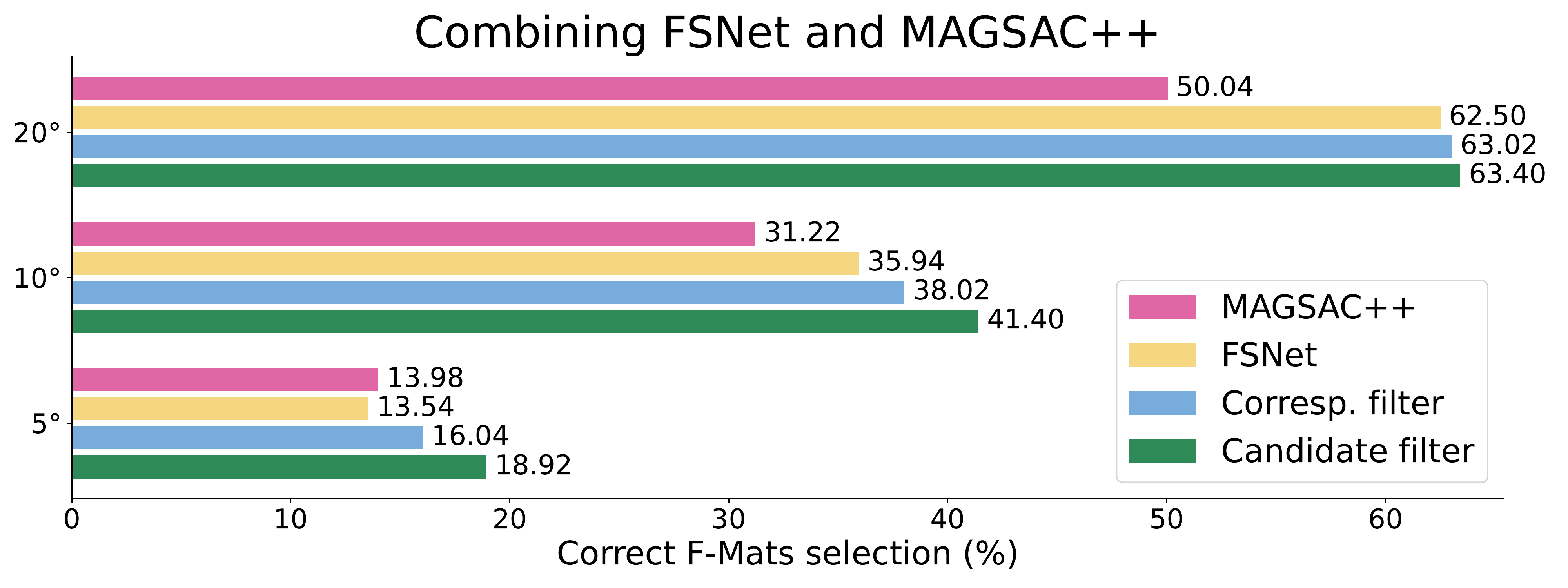}

   \caption{\textbf{Combining \ours and MAGSAC++.} 
   Complementing both scoring methods retrieves the highest number of correct \fmats under all error thresholds (5\textdegree, 10\textdegree ~and 20\textdegree).}
   \label{fig:fused_preds}
\end{figure}

%% file: tables/phototourism_table.tex
\begin{table}[t]
\footnotesize
\begin{center}
\begin{tabular}{c c l c l c}
\multicolumn{6}{c}{\textbf{PhotoTourism dataset}}\\ 
\cline{1-6}
\noalign{\smallskip}
\multicolumn{1}{c}{} & \multicolumn{3}{c}{\textbf{\maa} $\uparrow$} & \multicolumn{1}{c}{} & \multicolumn{1}{c}{\textbf{\mediandeg} $\downarrow$}\\ 
\cline{2-4} \cline{6-6}
\noalign{\smallskip}
\multicolumn{1}{c}{} & \multicolumn{1}{c}{\textbf{0-100}} & \multicolumn{1}{c}{} & \multicolumn{1}{c}{\textbf{All}} & \multicolumn{1}{c}{} & \multicolumn{1}{c}{\textbf{All}}\\
\cline{2-2} \cline{4-4} \cline{6-6}
\noalign{\smallskip}
\multicolumn{1}{c}{} & \multicolumn{1}{c}{R / t } & \multicolumn{1}{c}{} & \multicolumn{1}{c}{R / t}  & \multicolumn{1}{c}{} & \multicolumn{1}{c}{$e_R$ / $e_t$}\\ 
\hline \noalign{\smallskip}
Fundamental &  &&  &&  \\

\cline{1-1}
\noalign{\smallskip}
RANSAC \cite{fischler1981random} & 0.12 / 0.11 && 0.55 / 0.21 && 3.72 / 17.75 \\
MAGSAC++ \cite{barath2020magsac++} & \underline{0.24} / 0.17 && \underline{0.80} / \underline{0.44} && \underline{1.39} / \underline{5.42} \\
\ours (F) & 0.21 / \underline{0.19} && 0.71 / 0.35 && 2.08 / 7.22 \\
\hdashline\noalign{\smallskip}
w/ Corresp. filter & 0.21 / \underline{0.19} && \underline{0.80} / \underline{0.44} && \underline{1.39} / \underline{5.42} \\
w/ Candidate filter & \textbf{0.27} / \textbf{0.21} && \textbf{0.82} / \textbf{0.47} && \textbf{1.32} / \textbf{4.78} \\

\noalign{\smallskip}
\hline 
\hline \noalign{\smallskip}
Essential &  &&  &&  \\
\cline{1-1}
\noalign{\smallskip}
RANSAC \cite{fischler1981random} & 0.25 / \underline{0.31} && 0.81 / 0.59 && 1.13 / 3.06 \\
MAGSAC++ \cite{barath2020magsac++} & 0.24 / 0.30 && \underline{0.85} / \underline{0.65} && \underline{0.92} / \underline{2.18} \\
\ours (E) & \textbf{0.30} / \underline{0.31} && 0.81 / 0.55 && 1.39 / 3.42 \\
\hdashline\noalign{\smallskip}
w/ Corresp. filter & \textbf{0.30} / \underline{0.31} && \underline{0.85} / \underline{0.65} && \underline{0.92} / \underline{2.18} \\
w/ Candidate filter & \underline{0.29} / \textbf{0.35} && \textbf{0.87} / \textbf{0.67} && \textbf{0.88} / \textbf{2.08} \\
\noalign{\smallskip}
\end{tabular}
\end{center}
\vspace{-1.5em}
\normalsize
\caption{\textbf{Fundamental and essential matrix estimation on PhotoTourism dataset}. 
We show the mAA (10\textdegree) and median error when using SP-SG correspondences. Contrary to indoor datasets \cite{dai2017scannet}, in the outdoor scenario, current feature extractors provide very accurate and robust features, and in the test split only 1.5\% of the image pairs returned fewer than 100 correspondences. Nevertheless, combinations of \ours  and MAGSAC++ provide the most reliable methods for fundamental and essential matrix estimation. 
}
\label{tab:outdoor_tablev2}
\end{table}

%% file: tables/ablation_loss.tex

\begin{table}[t]
\footnotesize
\begin{center}
\begin{tabular}{c c l c l c}
\multicolumn{1}{c}{} & \multicolumn{3}{c}{\textbf{\maa} $\uparrow$} & \multicolumn{1}{c}{} & \multicolumn{1}{c}{\textbf{\mediandeg} $\downarrow$}\\ 
\cline{2-4} \cline{6-6}
\noalign{\smallskip}
\multicolumn{1}{c}{} & \multicolumn{1}{c}{\textbf{0-100}} & \multicolumn{1}{c}{} & \multicolumn{1}{c}{\textbf{All}} & \multicolumn{1}{c}{} & \multicolumn{1}{c}{\textbf{All}}\\
\cline{2-2} \cline{4-4} \cline{6-6}
\noalign{\smallskip}
\multicolumn{1}{c}{} & \multicolumn{1}{c}{R / t } & \multicolumn{1}{c}{} & \multicolumn{1}{c}{R / t}  & \multicolumn{1}{c}{} & \multicolumn{1}{c}{$e_R$ / $e_t$}\\ 
\hline \noalign{\smallskip}
Supervision &  & &   & &  \\
\cline{1-1}
\noalign{\smallskip}
RE1 & 0.23 / 0.12 &&  0.28 / 0.14  && 8.53 / 19.42 \\
CE as in \cite{barath2022learning} & 0.27 / \underline{0.18} &&  0.34 / \underline{0.20}  && 6.94 / \underline{13.21} \\
$L1$ & 0.25 / 0.16 && 0.32 / 0.18 && 7.52 / 14.07 \\
\ours (Soft-$L1$) & \textbf{0.29} / \textbf{0.19} && \underline{0.36} / \textbf{0.21} && \underline{6.52} / \textbf{12.05} \\
\hdashline\noalign{\smallskip}
MAGSAC++ \cite{barath2020magsac++} & \underline{0.28} / 0.12 &&  \textbf{0.38} / 0.17  && \textbf{6.35} / 17.89 \\
\end{tabular}
\end{center}
\normalsize
\vspace{-1.0em}
\caption{\textbf{Loss function and generalization ablation} results on the fundamental matrix estimation task in the test set of ScanNet. Supervising the training of \ours directly with the camera pose error provides the best results, while the CE approach outperforms L1 norm and Sampson distance (RE1) prediction models.
}
\vspace{-3pt}
\label{tab:loss_ablation}
\end{table}


%% file: figures/failures_plot.tex
\begin{figure}[t]
  \centering
   \includegraphics[width=0.96\linewidth]{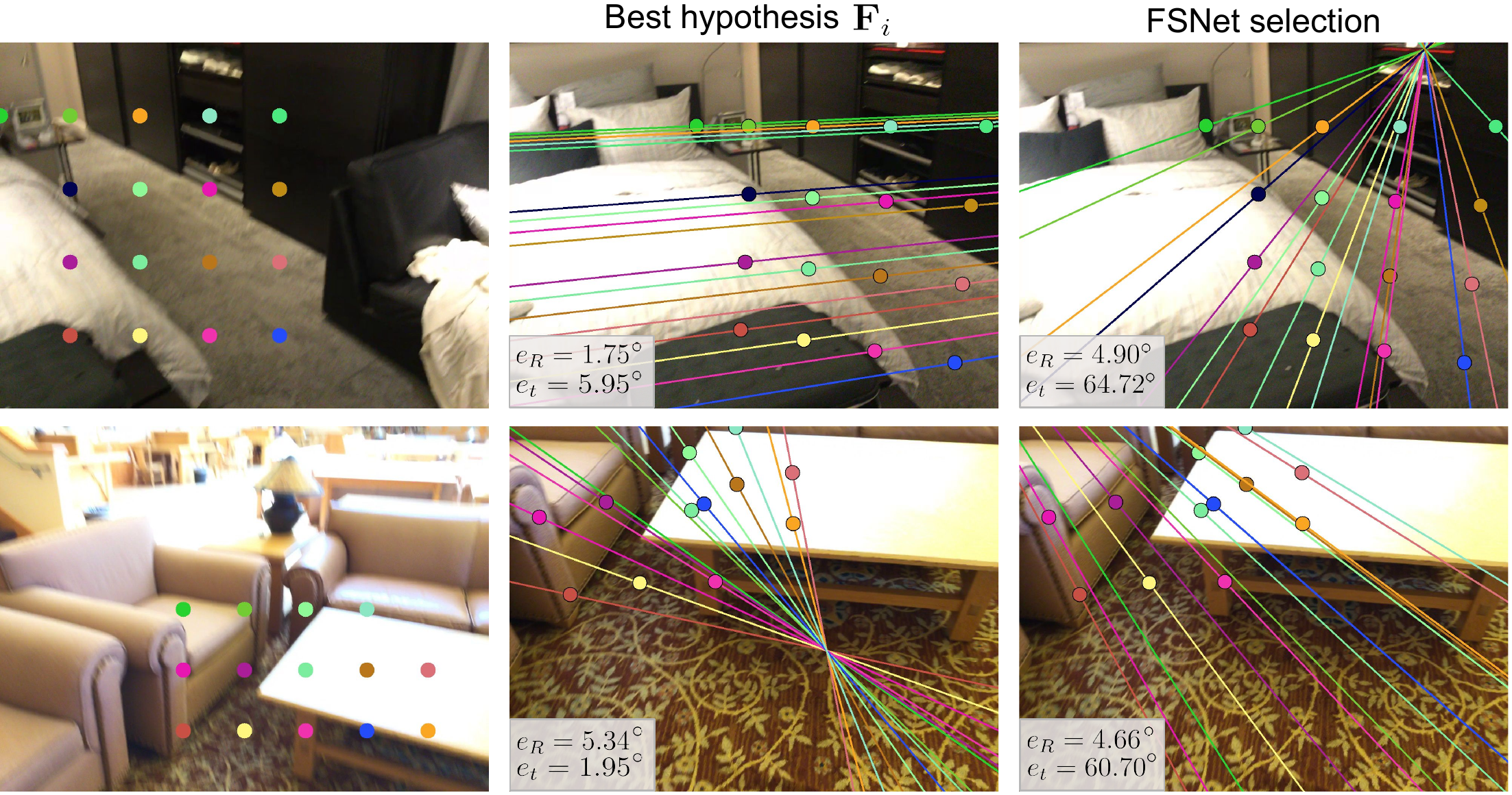}
   \caption{\textbf{Failure examples.} We plot the co-visible points and corresponding epipolar lines of the \fmat selected by \ours and the best hypothesis in the pool.
   Besides, we also mark the intersection of the best and \ours epipolar lines. 
   We see that the lines intersect at positions close to neighbor regions of the corresponding points, and hence, the low-resolution design of \ours limits its capability to select the optimal hypothesis.
   }
   \label{fig:failures}
   \vspace{-10pt}
\end{figure}


%% file: tables/supp_table_architecture.tex
\begin{table}[t]
\footnotesize
\begin{center}
\begin{tabular}{c c c}
\multicolumn{3}{c}{\textbf{Feature Extractor}}\\
\hline \noalign{\smallskip}
Layer & Description & Output Shape\\
\hline \noalign{\smallskip}
& Input Image & [b, 3, 256, 256]\\
0 & Conv-BN-ReLU & [b, 128, 256, 256]\\
1 & ResNet block 1 & [b, 128, 128, 128] \\
2 & ResNet block 2 & [b, 196, 64, 64] \\
3 & ResNet block 3 & [b, 256, 32, 32] \\
4 & ResNet block 4 & [b, 256, 16, 16] \\
5 & Up and Skip conn. w/ layer 3 & [b, 256, 32, 32] \\
6 & Conv-BN-LeakyReLU & [b, 196, 32, 32] \\
7 & Up and Skip conn. w/ layer 2 & [b, 196, 64, 64] \\
8 & Conv-BN-LeakyReLU & [b, 128, 64, 64] \\
\end{tabular}
\end{center}
\normalsize
\caption{\textbf{Feature extractor architecture details.} The feature extractor computes features from input images $A$ and $B$ at 1/4 of the input resolution. A ResNet block refers to a ResNet-18 \cite{he2016deep} block, which is composed of $3\times3$ convolutions, batch normalization layers \cite{ioffe2015batch}, ReLU activations \cite{agarap2018deep}, and a residual connection. The residual connection is done between the input to the block and the output. The Up and Skip conn. refers to an upsampling layer with bilinear interpolation and a skip connection between the input to the layer and the previous layer $i$.}
\label{tab:extractor}
\end{table}

\begin{table}[t]
\footnotesize
\begin{center}
\begin{tabular}{c c c }
\multicolumn{3}{c}{\textbf{Pose Error Regressor}}\\
\hline \noalign{\smallskip}
Layer & Description & Output Shape\\
\hline \noalign{\smallskip}
 & Input feature maps ($\mathbf{f}_i^A$ and $\mathbf{f}_i^B$) & [b, 128, 32, 32] \\
1 & ResNet block 1 & [b, 128, 16, 16] \\
2 & ResNet block 2 & [b, 128, 8, 8] \\
3 & ResNet block 3 & [b, 256, 4, 4] \\
4 & ResNet block 4 & [b, 512, 2, 2] \\
5 & 2D Avg. Pooling ($\mathbf{v}_i^{A\rightarrow B}$ and $\mathbf{v}_i^{B \rightarrow A}$) & [b, 512, 1, 1] \\
6 & Max Pooling ($\mathbf{v}_i$) & [b, 512, 1, 1] \\
7 & Conv1x1-BN-ReLU (MLP layer 1) & [b, 512, 1, 1] \\
8 & Conv1x1-BN-ReLU (MLP layer 2) & [b, 256, 1, 1] \\
9 & Conv1x1-BN-ReLU (MLP layer 3) & [b, 2] \\
\end{tabular}
\end{center}
\normalsize
\caption{\textbf{Pose error regressor architecture details.} The pose error regressor block estimates the rotation ($e^R_i$) and the translation ($e^t_i$) errors for images $A$ and $B$ and fundamental matrix $F_i$. The input to the pose error regressor block is the epipolar transformed features $\mathbf{f}_i^A$ and $\mathbf{f}_i^B$. As in the feature extractor, the ResNet block refers to a ResNet-18 \cite{he2016deep} block.}
\label{tab:arch_err_regressor}
\end{table}

%% file: tables/loftr_table.tex
\begin{table}[t]
\footnotesize
\begin{center}
\begin{tabular}{c c l c}
\multicolumn{1}{c}{} & \multicolumn{1}{c}{\textbf{\maa} $\uparrow$} & \multicolumn{1}{c}{} & \multicolumn{1}{c}{\textbf{\mediandeg} $\downarrow$}\\ 
\cline{2-2} \cline{4-4}
\noalign{\smallskip}
\multicolumn{1}{c}{} & \multicolumn{1}{c}{R / t / max(R,t)}  & \multicolumn{1}{c}{} & \multicolumn{1}{c}{$e_R$ / $e_t$}\\ 
\hline \noalign{\smallskip}
\noalign{\smallskip}
Fundamental & &&\\
\cline{1-1}
\noalign{\smallskip}
\cline{1-1}
\noalign{\smallskip}


\textbf{LoFTR}  \cite{sun2021loftr} & && \\
\cline{1-1}
\noalign{\smallskip}
MAGSAC++ \cite{barath2020magsac++} & 0.11 / 0.03 / 0.02 && 27.21 / 46.53 \\
\ours & \textbf{0.13} / \textbf{0.05} / \textbf{0.03} && \textbf{20.98} / \textbf{38.42} \\
w/ Corresp. filter & 0.11 / 0.03 / 0.02 && 26.48 / 46.28 \\
w/ Candidate filter & \textbf{0.13} / 0.04 / \textbf{0.03} && 21.98 / 40.34 \\

\hdashline\noalign{\smallskip}

\textbf{SIFT}  \cite{lowe1999object} & && \\
\cline{1-1}
\noalign{\smallskip}
MAGSAC++ \cite{barath2020magsac++} & 0.08 / 0.03 / 0.02 && 87.26 / 50.02 \\
\ours & 0.08 / 0.03 / 0.02 && \textbf{36.78} / \textbf{43.96} \\
w/ Corresp. filter & 0.08 / 0.03 / 0.02 && 47.21 / 47.94 \\
w/ Candidate filter & \textbf{0.09} / 0.03 / 0.02 && 45.35 / 46.47 \\

\hline
\hline \noalign{\smallskip}
Essential & &&\\
\cline{1-1}
\noalign{\smallskip}
\cline{1-1}
\noalign{\smallskip}


\textbf{LoFTR}  \cite{sun2021loftr} & &&\\
\cline{1-1}
\noalign{\smallskip}
MAGSAC++ \cite{barath2020magsac++} & 0.17 / 0.09 / \textbf{0.07} && 22.50 / 37.79 \\
\ours & \textbf{0.20} / 0.10 / 0.07 && \textbf{17.56} / \textbf{31.79} \\
w/ Corresp. filter & 0.17 / 0.09 / 0.07 && 21.98 / 37.64 \\
w/ Candidate filter & \textbf{0.20} / \textbf{0.11} / \textbf{0.08} && 18.88 / 34.02 \\

\hdashline\noalign{\smallskip}

\textbf{SIFT}  \cite{lowe1999object} & && \\
\cline{1-1}
\noalign{\smallskip}
MAGSAC++ \cite{barath2020magsac++} & 0.14 / 0.06 / 0.05 && 73.28 / 46.06 \\
\ours & 0.15 / 0.07 / \textbf{0.06} && \textbf{27.73} / \textbf{38.74} \\
w/ Corresp. filter & 0.15 / 0.06 / 0.05 && 42.03 / 43.11 \\
w/ Candidate filter & \textbf{0.16} / \textbf{0.08} / \textbf{0.06} && 38.05 / 41.38 \\

\end{tabular}
\end{center}
\vspace{-1em}
\normalsize
\caption{\textbf{Integrating \ours with LoFTR \cite{sun2021loftr} and SIFT \cite{lowe1999object}.} \maa and Median error (\textdegree\xspace) results for \ours and MAGSAC++ on the fundamental and essential matrix estimation task on the ScanNet indoor dataset. 
As a reference, LoFTR detects fewer than 100 correspondences on 6.5\% of the image pairs. 
Meanwhile, the test set based on SIFT correspondences
results in 3,319 (0-100) and 1,681 (100-Inf) image pairs. We observe that when there are lower quality of correspondences, \ours comes out ahead, \textit{i.e.}, \ours (alone) returns the lowest pose errors.  
}
\label{tab:loftr_results}
\end{table}

%% file: tables/SP_more_corr.tex
\begin{table}[t]
\footnotesize
\begin{center}
\begin{tabular}{c c l c}
\multicolumn{1}{c}{} & \multicolumn{1}{c}{\textbf{\maa} $\uparrow$} & \multicolumn{1}{c}{} & \multicolumn{1}{c}{\textbf{\mediandeg} $\downarrow$}\\ 
\cline{2-2} \cline{4-4}
\noalign{\smallskip}
\multicolumn{1}{c}{} & \multicolumn{1}{c}{R / t / max(R,t)}  & \multicolumn{1}{c}{} & \multicolumn{1}{c}{$e_R$ / $e_t$}\\ 
\hline \noalign{\smallskip}
\noalign{\smallskip}
Fundamental & &&\\
\cline{1-1}
\noalign{\smallskip}
\textbf{SP-SG*} + MAGSAC++ & 0.38 / 0.18 / 0.14 && 6.22 / 16.50 \\
\hdashline\noalign{\smallskip}
SP-SG + MAGSAC++ & 0.38 / 0.17 / 0.14 && 6.35 / 17.89 \\
\ours & 0.36 / 0.21 / 0.15 && 6.52 / 12.05 \\
w/ Corresp. filter  & 0.39 / 0.22 / 0.16 && 6.09 / 11.59 \\
w/ Candidate filter  & \textbf{0.43} / \textbf{0.23} / \textbf{0.19} && \textbf{5.38} / \textbf{11.39} \\
\hline
\hline \noalign{\smallskip}
Essential & &&\\
\cline{1-1}
\noalign{\smallskip}
\textbf{SP-SG*} + MAGSAC++ & 0.38 / 0.24 / 0.19 && 6.38 / 11.30 \\
\hdashline\noalign{\smallskip}
SP-SG + MAGSAC++ & 0.40 / 0.26 / 0.21 && 5.95 / 10.96 \\
\ours & \textbf{0.44} / 0.28 / 0.22 && 5.13 / 8.95 \\
w/ Corresp. filter  & \textbf{0.44} / \textbf{0.29} / \textbf{0.23} && \textbf{5.07} / \textbf{8.75} \\
w/ Candidate filter  & \textbf{0.44} / 0.28 / \textbf{0.23} && 5.14 / 9.48 \\
\end{tabular}
\end{center}
\vspace{-1em}
\normalsize
\caption{\textbf{SuperGlue with more correspondences in the indoor ScanNet dataset}. 
SP-SG* refers to SP-SG with dynamic matching confidence threshold, which is reduced in order to either obtain at least 100 correspondences or the maximum correspondences that SP-SG provides. It can be seen that naively increasing the number of correspondences does not lead to improved results. 
}
\label{tab:SP_more_corr}
\end{table}

%% file: figures/supp_F_E_dist.tex
\begin{figure*}[t]
  \centering
   \includegraphics[width=1.\linewidth]{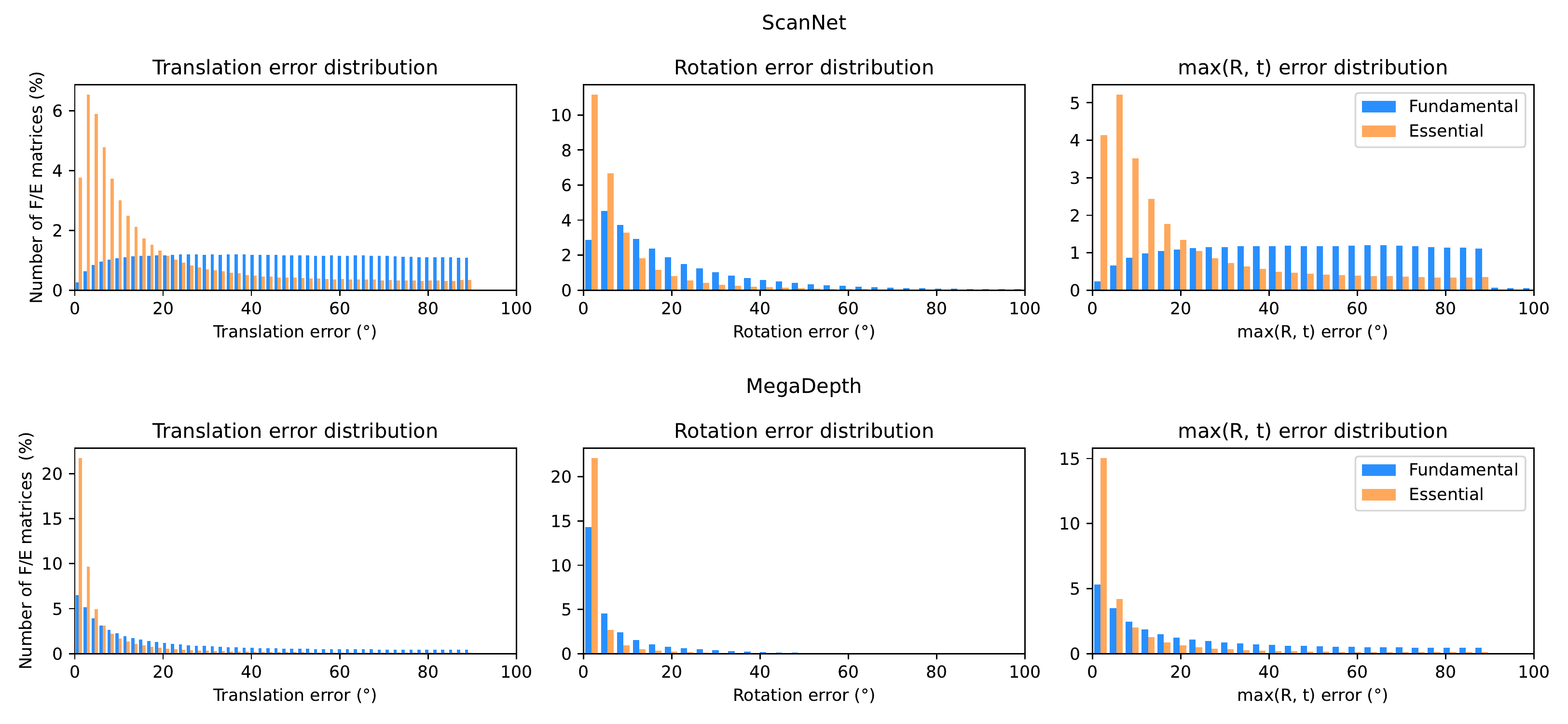}
   \caption{\textbf{Fundamental vs essential matrix error distributions.} The figure shows the error distributions (\textdegree), translation, rotation, and the maximum of both, for the generated fundamental and essential matrices with SP-SG correspondences. We observe that essential matrices have a higher population on the low error regime, \textit{i.e.}, matrices with pose error below 20\textdegree. Meanwhile, fundamental matrices show a wider range of errors, especially in the indoor scenario, where correspondences do not provide enough reliability for accurate two-view geometry estimation. This observation leads us to train \ours for either the task of fundamental or essential matrix estimation. } 
   \label{fig:supp_F_vs_E_dist}
\end{figure*}

%% file: tables/SuppMat_F_E_ScanNet.tex
\begin{table*}[ht]
\footnotesize
\begin{center}
\hspace{-1em}
\begin{tabular}{c c l c  l c  l c l c  l  c}
\multicolumn{1}{c}{} & \multicolumn{3}{c}{\textbf{0-100}} & \multicolumn{1}{c}{} & \multicolumn{3}{c}{\bf{100-Inf}} & \multicolumn{1}{c}{} & \multicolumn{3}{c}{\bf{All}}\\ 
\cline{2-4}
\cline{6-8}
\cline{10-12}
\noalign{\smallskip}
\multicolumn{1}{c}{} & \multicolumn{1}{c}{\textbf{\maa} $\uparrow$} & \multicolumn{1}{c}{} & \multicolumn{1}{c}{\textbf{\mediandeg} $\downarrow$} & \multicolumn{1}{c}{} & \multicolumn{1}{c}{\textbf{\maa} $\uparrow$} & \multicolumn{1}{c}{} & \multicolumn{1}{c}{\textbf{\mediandeg} $\downarrow$} & \multicolumn{1}{c}{} & \multicolumn{1}{c}{\textbf{\maa} $\uparrow$} & \multicolumn{1}{c}{} & \multicolumn{1}{c}{\textbf{\mediandeg} $\downarrow$}\\ 
\cline{2-2}
\cline{4-4}
\cline{6-6}
\cline{8-8}
\cline{10-10}
\cline{12-12}
\noalign{\smallskip}
 & R / t / max(R, t) && $e_R$ / $e_t$ && R / t / max(R, t)  && $e_R$ / $e_t$ && R / t / max(R, t)  && $e_R$ / $e_t$\\
\hline \noalign{\smallskip}
Fundamental & &&  &&  &&  &&  && \\
\cline{1-1}
\noalign{\smallskip}
MAGSAC++ \cite{barath2020magsac++} & 0.41 / 0.17 / 0.14 && 5.55 / 16.44 && 0.76 / 0.33 / 0.32 && 1.77 / 7.74 && 0.61 / 0.26 / 0.24 && 2.71 / 10.49  \\
\ours (F) & 0.41 / 0.22 / 0.17 && 5.62 / 10.84 && 0.62 / 0.24 / 2.22 && 3.02 / 11.07 && 0.53 / 0.23 / 0.19 && 3.87 / 10.91 \\
\hdashline\noalign{\smallskip}
w/ Corresp. filter & 0.41 / 0.22 / 0.17 && 5.62 / 10.84 && 0.76 / 0.33 / 0.32 && \textbf{1.77} / 7.74 && 0.61 / 0.28 / 0.25 && 2.82 / 8.91 \\
w/ Candidate filter  & \textbf{0.49} / \textbf{0.26} / \textbf{0.21} && \textbf{4.36} / \textbf{10.47} && \textbf{0.78} / \textbf{0.38} / \textbf{0.36} && 1.79 / \textbf{6.06} && \textbf{0.65} / \textbf{0.32} / \textbf{0.30} && \textbf{2.46} / \textbf{7.58}\\
\hline
\hline \noalign{\smallskip}
Essential & &&  &&  &&  &&  && \\
\cline{1-1}
\noalign{\smallskip}
MAGSAC++ \cite{barath2020magsac++} & 0.42 / 0.26 / 0.21 && 5.39 / 10.96 && 0.76 / 0.41 / 0.40 && 1.72 / 5.58 && 0.61 / 0.34 / 0.31 && 2.63 / 7.41 \\
\ours (E) & 0.47 / 0.28 / 0.22 && 4.57 / \textbf{8.85} && 0.72 / 0.35 / 0.33 && 2.22 / 6.80 && 0.61 / 0.32 / 0.28 && 3.06 / 7.53 \\
\hdashline\noalign{\smallskip}
w/ Corresp. filter & 0.47 / 0.28 / 0.22 && 4.57 / \textbf{8.85} && 0.76 / 0.41 / 0.40 && 1.72 / 5.58 && 0.64 / 0.35 / 0.32 && 2.66 / 6.82\\
w/ Candidate filter  & \textbf{0.48} / \textbf{0.29} / \textbf{0.24} && \textbf{4.47} / 9.26 && \textbf{0.79} / \textbf{0.45} / \textbf{0.43} && \textbf{1.69} / \textbf{4.97} && \textbf{0.66} / \textbf{0.38} / \textbf{0.35} && \textbf{2.43} / \textbf{6.36}\\
\end{tabular}
\end{center}
\normalsize
\caption{\textbf{Fundamental and essential matrix estimation on SuperGlue's \cite{sarlin2020superglue} test set of ScanNet}.
We compute the \maa and \mediandeg metrics in the SuperGlue's test split of ScanNet, and divide the pairs of images based on the number of SP-SG correspondences. The split results in 649 (0-100) and 851 (100-Inf) image pairs. We see that when the number of correspondences is small (0-100), \ours outperforms MAGSAC++ in both, fundamental and essential matrix estimation. In the overall split (All), the best results are obtained when combining \ours and MAGSAC++ hypothesis scores.
}
\label{tab:sup_F_E_estimation_table}
\end{table*}

%% file: figures/overlap_dist_tests.tex
\begin{figure}[t]
  \centering
   \includegraphics[width=1.0\linewidth]{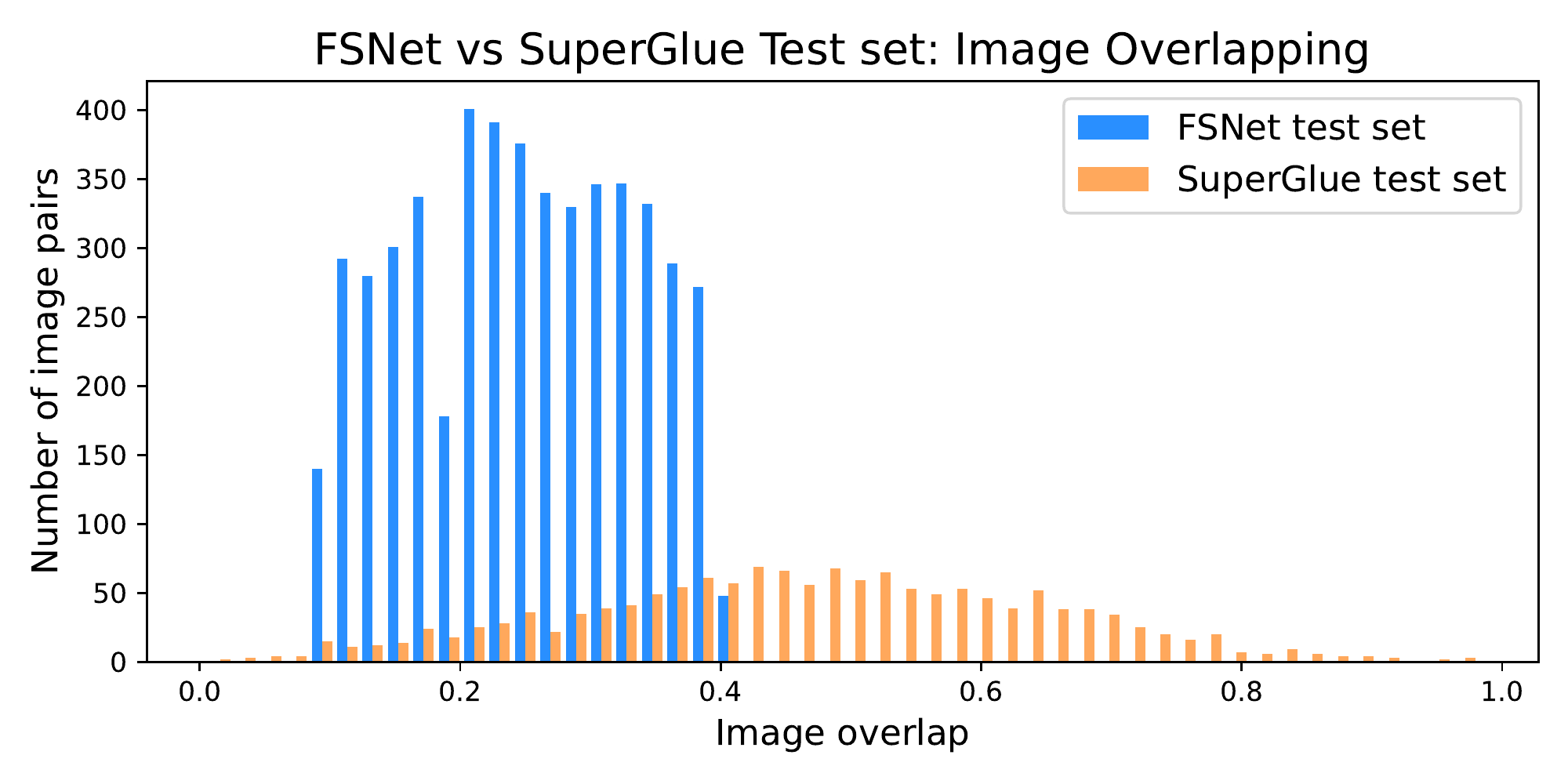}
   \caption{\textbf{Overlapping distribution of the image pairs of \ours and SuperGlue \cite{sarlin2020superglue} test sets.}
   \ours focuses on image pairs where current correspondence-scoring methods struggle. 
    Therefore, to tackle such cases, we generate \ours training, validation, and test sets with low visual overlapping images, where correspondences are few and then, in many cases, not reliable. 
   }
   \label{fig:overlap_dist_test}
\end{figure}

%% file: figures/candidate_filter.tex
\begin{figure}
     \centering
     \begin{subfigure}[b]{0.48\textwidth}
         \centering
         \includegraphics[width=\textwidth]{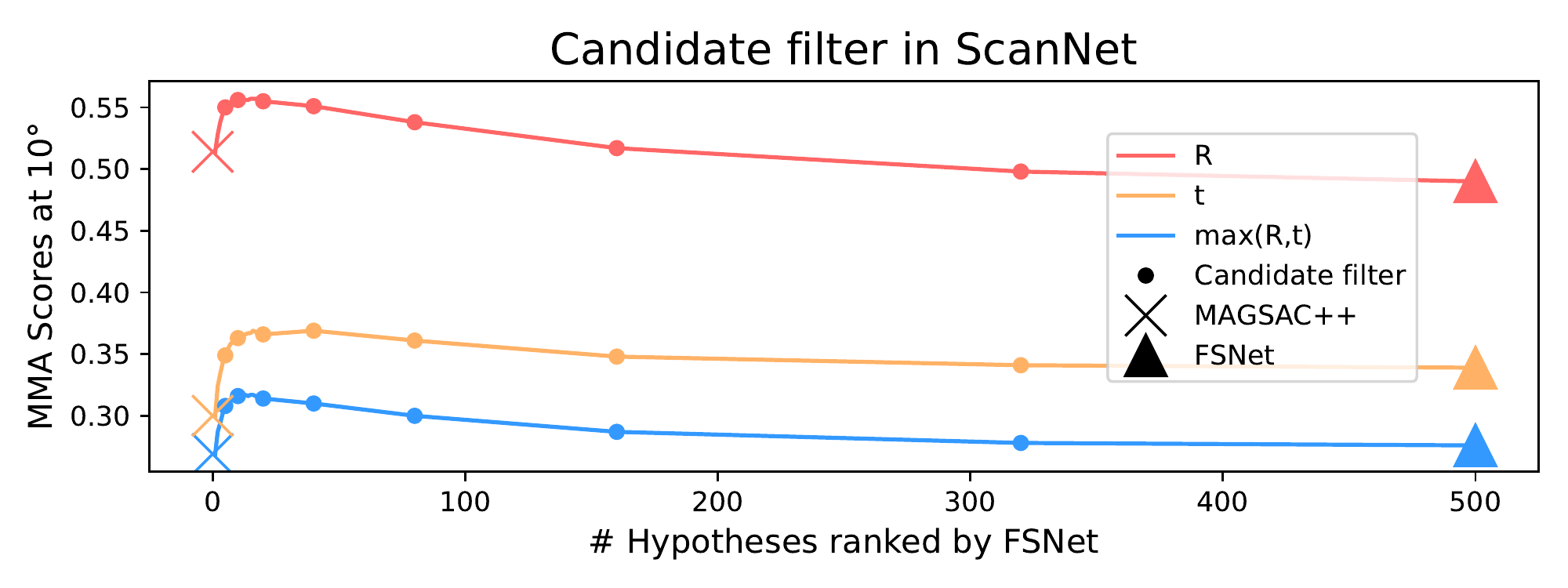}
         \caption{FSNet with candidate filter in ScanNet validation set.}
     \end{subfigure}
     \hfill
     \vspace{0.3em}
     \begin{subfigure}[b]{0.48\textwidth}
         \centering
         \includegraphics[width=\textwidth]{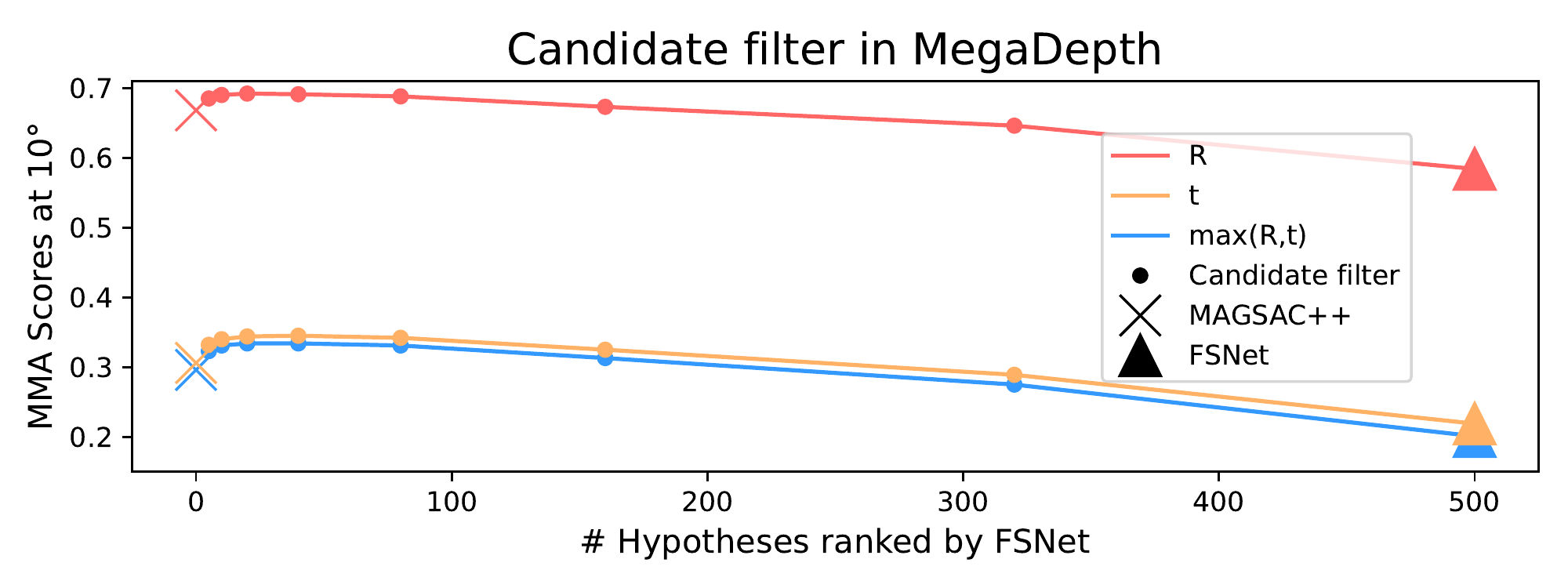}
         \caption{FSNet with candidate filter in MegaDepth validation set.}
     \end{subfigure}
        \caption{\textbf{Combination of \ours and MAGSAC++} based on the proposed candidate filter in the fundamental matrix estimation task. 
        When using the candidate filter, \ours only scores the top-k MAGSAC++ hypotheses. 
        If the number of hypotheses to score is 1, it refers to original MAGSAC++, while 500 hypotheses corresponds to \ours alone. As a reference, we also indicate MAGSAC++ ($\times$) and \ours ($\blacktriangle$) scores.}
        \label{fig:candidate_filter}
\end{figure}

%% file: tables/unimodal_vs_multimodal.tex
\begin{table*}[ht!]
\footnotesize
\begin{center}
\hspace{-1em}
\begin{tabular}{c c l c  l c  l c l c  l  c}
\multicolumn{1}{c}{} & \multicolumn{3}{c}{\textbf{Unimodal} (39.16\%)} & \multicolumn{1}{c}{} & \multicolumn{7}{c}{\textbf{Multimodal} (60.84\%)}\\ 
\cline{2-4}
\cline{6-12}
\noalign{\smallskip}
 & FSNet && MAGSAC++ && FSNet ($e < 10^{\circ}$) && MAGSAC++ ($e < 10^{\circ}$) && Both with $e < 10^{\circ}$  && Both with $e \geq 10^{\circ}$ \\
\hline \noalign{\smallskip}
\% & 69.40 && 30.60 && 14.34 && 3.55 && 16.67 && 65.43
\end{tabular}
\end{center}
\normalsize
\caption{\textbf{Distribution of top ranking hypotheses by MAGSAC++ scores in ScanNet validation set}.
We report the percentage of times we find an unimodal or multimodal distribution among the top 5 hypotheses returned by MAGSAC++. The criteria to determine if a pool of hypotheses is unimodal or multimodal is based on their pairwise distances. If the difference between the minimum and maximum distance is above 10$^\circ$, then we indicate it as a multimodal, otherwise, we mark the pool as unimodal. Besides the distinction between unimodal or multimodal, we also indicate which method was able to select the best hypothesis (unimodal), or which method was able to select a valid hypothesis (a hypothesis with a pose error below 10$^\circ$ \textit{w.r.t.} the ground-truth pose).
}
\label{tab:uni_vs_multi}
\end{table*}

%% file: figures/uni_vs_multi.tex
\begin{figure*}
     \centering
     \begin{subfigure}[b]{0.99\textwidth}
         \centering
         \includegraphics[width=\textwidth]{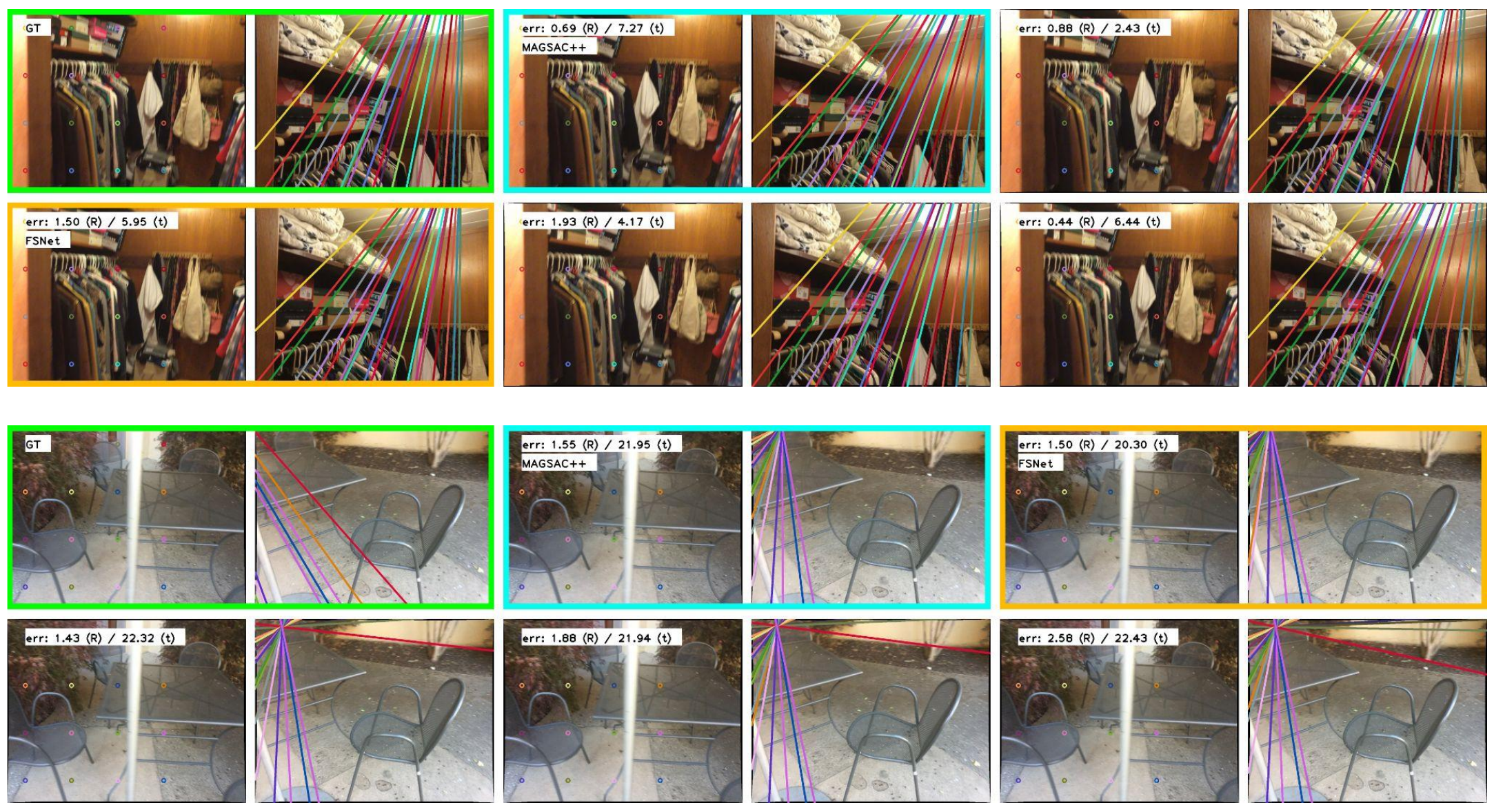}
         \caption{Unimodal distribution.}
     \end{subfigure}
     \hfill
     \vspace{2em}
     \begin{subfigure}[b]{0.99\textwidth}
         \centering
         \includegraphics[width=\textwidth]{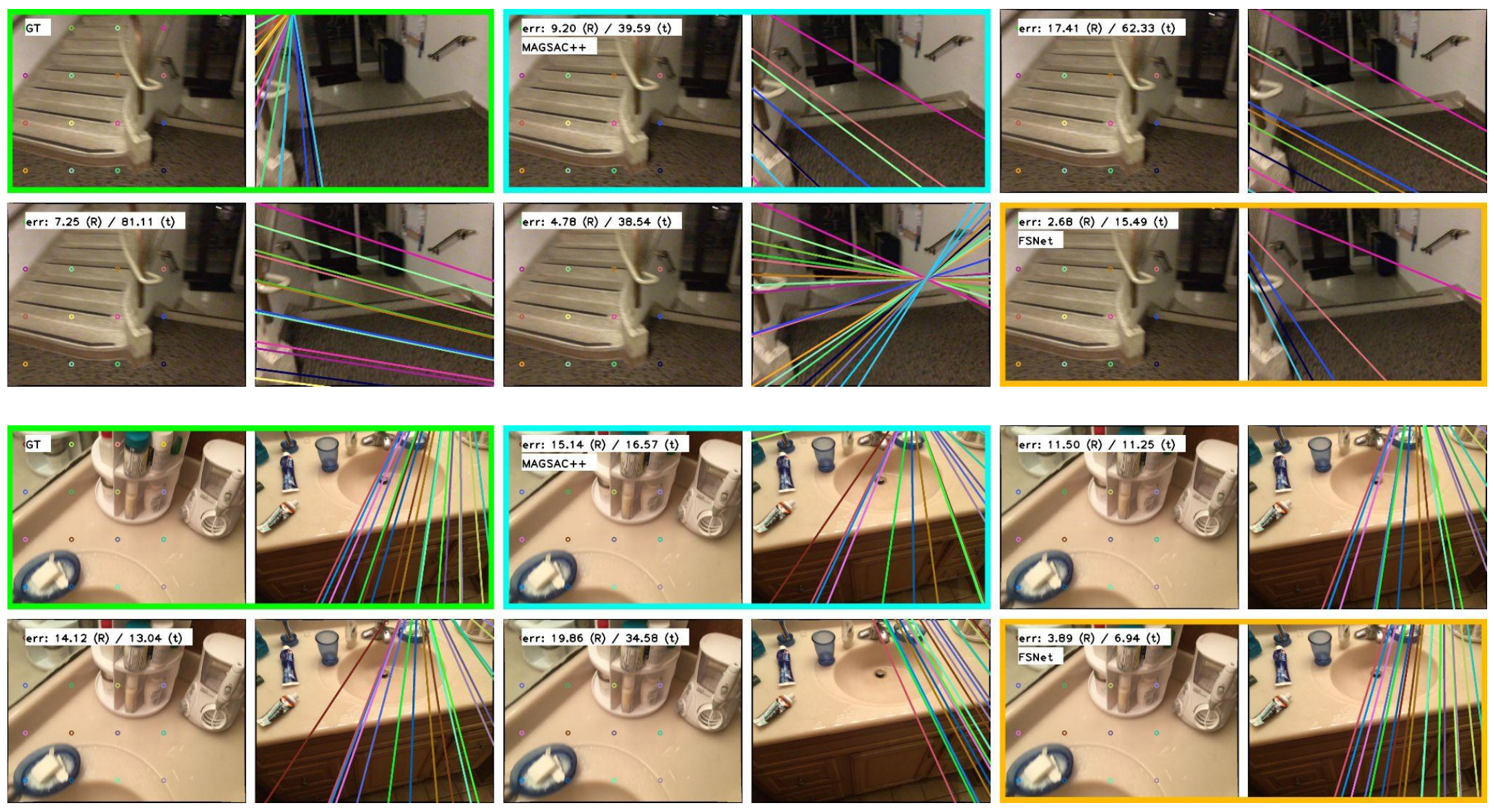}
         \caption{Multimodal distribution.}
     \end{subfigure}
        \caption{Example of top scoring hypotheses returned by MAGSAC++. (a) Shows unimodal examples, where all returned hypotheses are similar, while (b) returns hypotheses with different distributions. Ground-truth is boxed in green, MAGSAC++ selection in blue, and \ours in orange.}
        \label{fig:uni_vs_multi}
\end{figure*}